\documentclass[10pt,journal,compsoc]{IEEEtran}
\ifCLASSOPTIONcompsoc
  \usepackage[nocompress]{cite}
\else
  \usepackage{cite}
\fi
\ifCLASSINFOpdf
  \usepackage[pdftex]{graphicx}
\else
\fi
\usepackage{amsmath}
\usepackage{amssymb}

\usepackage{multirow}
\usepackage{booktabs,siunitx}
\usepackage[table,xcdraw]{xcolor}
\usepackage{graphicx}
\usepackage{diagbox}
\usepackage{ulem}
\usepackage{colortbl}
\usepackage{url}
\usepackage{hyperref}
\usepackage[switch]{lineno}

\begin{document}
% \title{Bare Demo of IEEEtran.cls for\\ IEEE Computer Society Journals}
\title{ControLRM: Fast and Controllable 3D Generation via Large Reconstruction Model}

\author{
% Hongbin~Xu, Weitao~Chen, Zhipeng~Zhou, Feng~Xiao, Baigui~Sun, Liefeng~Bo, Mike~Zheng~Shou, Wenxiong~Kang,
Hongbin~Xu, Weitao~Chen, Zhipeng~Zhou, Feng~Xiao, Baigui~Sun, Mike~Zheng~Shou, Wenxiong~Kang,
\IEEEcompsocitemizethanks{
\IEEEcompsocthanksitem H. Xu, F. Xiao and W. Kang are with  South China University of Technology, Guangzhou, China.
\IEEEcompsocthanksitem W. Chen and B. Sun are with Alibaba Group, Hangzhou, China.
\IEEEcompsocthanksitem Z. Zhou is with University of Chinese Academy of Sciences.
\IEEEcompsocthanksitem Mike Z. Shou is with National University of Singapore, Singapre.
\IEEEcompsocthanksitem W. Kang is the corresponding author. (E-mail: auwxkang@scut.edu.cn)
% \IEEEcompsocthanksitem Hongbin Xu, Feng Xiao and Wenxiong Kang are with the School of Automation Science and Engineering, South China University of Technology, Guangzhou, China, 510006.
% \IEEEcompsocthanksitem Weitao Chen, Baigui Sun, and Liefeng Bo are with Alibaba Group, Hangzhou, China.
% \IEEEcompsocthanksitem Mike Zheng Shou is with National University of Singapore, Singapre.
% \IEEEcompsocthanksitem Wenxiong Kang is the corresponding author. (E-mail: auwxkang@scut.edu.cn)
}
}


\markboth{Journal of \LaTeX\ Class Files,~Vol.~14, No.~8, August~2015}%
{Shell \MakeLowercase{\textit{et al.}}: Bare Demo of IEEEtran.cls for Computer Society Journals}
\IEEEtitleabstractindextext{%
\begin{abstract}
Despite recent advancements in 3D generation methods, achieving controllability still remains a challenging issue. 
Current approaches utilizing score-distillation sampling are hindered by laborious procedures that consume a significant amount of time. 
Furthermore, the process of first generating 2D representations and then mapping them to 3D lacks internal alignment between the two forms of representation.
To address these challenges, we introduce ControLRM, an end-to-end feed-forward model designed for rapid and controllable 3D generation using a large reconstruction model (LRM). 
ControLRM comprises a 2D condition generator, a condition encoding transformer, and a triplane decoder transformer. 
Instead of training our model from scratch, we advocate for a joint training framework. 
In the condition training branch, we lock the triplane decoder and reuses the deep and robust encoding layers pretrained with millions of 3D data in LRM.
In the image training branch, we unlock the triplane decoder to establish an implicit alignment between the 2D and 3D representations. 
To ensure unbiased evaluation, we curate evaluation samples from three distinct datasets (G-OBJ, GSO, ABO) rather than relying on cherry-picking manual generation.
The comprehensive experiments conducted on quantitative and qualitative comparisons of 3D controllability and generation quality demonstrate the strong generalization capacity of our proposed approach. 
For access to our project page and code, please visit \href{https://toughstonex.github.io/controlrm.github.io/}{our project page}.

% : \href{ControLRM Project Page}{https://toughstonex.github.io/controlrm}.
% Despite recent advancements in 3D generation methods, the topic of controllability is still a challenging problem.
% Existing works based on score-distillation sampling suffer from tedious procedures, consuming huge amount of time.
% Moreover, the tedious procedures of generating 2D first and map to 3D lacks an internal alignment between 2D and 3D representation.
% To handle these issues, we propose ControLRM, an end-to-end feed-forward model for fast and controllable 3D generation with large reconstruction model.
% ControLRM contains a 2D condition generator, a condition encoding transformer and a triplane decoder transformer.
% Instead of training our model from the scratch, we propose a joint training framework.
% In the condition training branch, we lock the triplane decoder and reuses the deep and robust encoding layers pretrained with millions of 3D data.
% In the image training branch, we unlock the triplane decoder and provide an implicit alignment between 2D and 3D representations.
% To avoid cherry-picking, we collect evaluation samples from 3 different datasets (G-OBJ, GSO, ABO) instead of manually generating them.
% The extensive experiments on the quantitative benchamrks and qualitative comparisons of 3D controllability and generation quality reveals the strong generalization ability of our proposed method.
% Our project page and code are available at: https://toughstonex.github.io/controlrm.
\end{abstract}

}

\maketitle

\IEEEdisplaynontitleabstractindextext

% \fi
\IEEEpeerreviewmaketitle

% ================================================================================
% 已用GPT润色
% ================================================================================

\IEEEraisesectionheading{\section{Introduction}\label{sec:introduction}}
\label{sec-introduction}

% 已用GPT润色
\IEEEPARstart{T}he potential of 3D content generation spans various sectors such as digital games, virtual reality/augmented reality (VR/AR), and filmmaking.
Fundamental techniques in 3D content creation, such as text-to-3D and image-to-3D methods, offer substantial benefits by significantly reducing the need for laborious and costly manual work among professional 3D artists, thus enabling individuals without expertise to engage in the creation of 3D assets.
Given the notable achievements in 2D content generation, exemplified by projects like DALL-E \cite{ramesh2021zero} and StableDiffusion \cite{rombach2022high}, the community is increasingly focusing on advancements in 3D content generation.
Recent progress in this field is credited to the advantageous characteristics of image diffusion models \cite{rombach2022high,liu2023zero}, differentiable 3D representations \cite{mildenhall2021nerf,kerbl20233d}, and large reconstruction models \cite{hong2023lrm,tang2024lgm}.

% 已用GPT润色
An appealing area of interest for 3D content creation is \textbf{text-to-3D} generation. 
Some groundbreaking advancements \cite{poole2022dreamfusion,chen2023fantasia3d} in text-to-3D synthesis have introduced methods to enhance a neural radiance field (NeRF) \cite{mildenhall2021nerf} through score distillation sampling (SDS) loss \cite{poole2022dreamfusion} for 3D asset generation. 
Building upon the influential work of DreamFusion \cite{poole2022dreamfusion}, these SDS-based techniques aim to distill 3D information from pretrained large text-to-image generative models \cite{ramesh2021zero,rombach2022high}. 
Various strategies seek to elevate generation quality by expanding to multiple optimization phases \cite{chen2023fantasia3d}, optimizing 3D representation and diffusion prior simultaneously \cite{sun2023dreamcraft3d,wang2024prolificdreamer}, and adjusting score distillation algorithms \cite{katzir2023noise,yu2023text}.

\begin{figure*}[t]
\centering
\includegraphics[width=\linewidth]{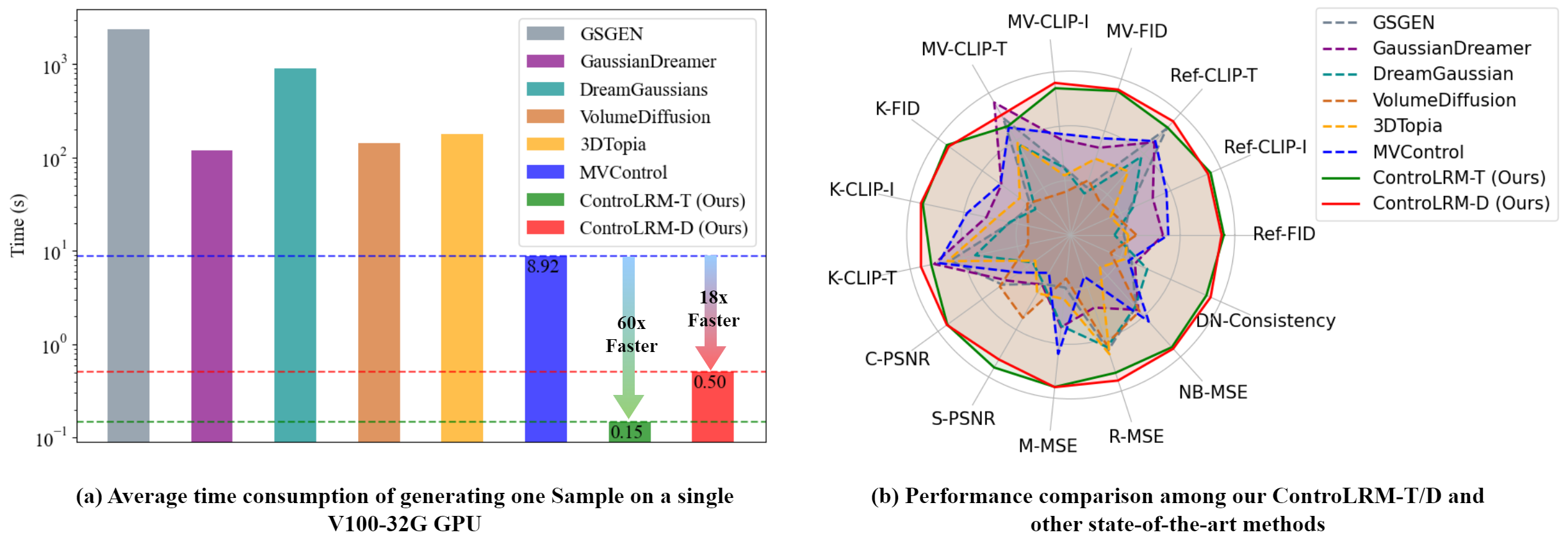}
\vspace{-0.7cm}
\caption{Performance and efficiency comparison among different conditional 3D generation methods. 
Fig. (a) shows the average time consumption on a single V100-32G GPU of different methods. Our ControLRM-T and ControLRM-D can respectively achieve 60 and 18 times faster inference speed compared with the fastest baseline, MVControl\cite{li2024controllable}. Fig (b) shows the results of 15 evaluation metrics on the G-Objaverse test set, including 3D controllability metrics (introduced in Sec. \ref{exp:controllability:metrics}) and controllable 3D generation metrics (introduced in Sec. \ref{exp:generation:metrics}).}
\vspace{-0.4cm}
\label{fig:introduction}
\end{figure*}

% 已用GPT润色
Another crucial aspect of generating 3D content is the process of \textbf{image-to-3D} synthesis.
The traditional approach to this challenge relies on 3D reconstruction methods such as Structure-from-Motion \cite{schonberger2016structure} and Multi-view Stereo \cite{goesele2007multi,yao2018mvsnet,xu2021self,xu2021digging}. 
These techniques involve identifying 3D surface points by comparing similarities among point features extracted from source images, enabling the creation of highly precise surface and texture maps. 
Despite significant achievements in accurately reconstructing geometrical details, these methods still struggle to reproduce detailed view-dependent appearances. 
Consequently, recent advancements have focused on developing implicit 3D representations like neural radiance fields \cite{mildenhall2021nerf,kerbl3Dgaussians} and neural implicit surfaces \cite{niemeyer2020differentiable,yariv2020multiview}. 
These novel approaches explore volumetric representations that can be learned from dense multi-view datasets without explicit feature matching, offering more efficient and high-quality solutions \cite{barron2021mip,chen2022tensorf,kerbl3Dgaussians}. 
Such efforts aim to move towards feed-forward models for radiance fields reconstruction, relaxing the need for dense views and per-scene optimization. 
Leveraging the capabilities and generalization power of large generative models like diffusion models, recent studies \cite{liu2023zero1to3,li2023sweetdreamer,liu2023syncdreamer,long2024wonder3d,shi2023mvdream} have integrated pre-trained generative models with multi-view information to generate new views from sparse inputs. 
Additionally, the emergence of Large Reconstruction Models (LRM) \cite{hong2023lrm,tochilkin2024triposr,zou2024triplane} has emphasized learning internal perspective relationships through a triplane transformer \cite{chan2022efficient} and cross-attention mechanisms with 2D visual features from single-view input images. 
Recent enhancements \cite{tang2024lgm,zhang2024gs} of LRM have focused on replacing triplane-based volume rendering with 3D Gaussian splatting \cite{kerbl3Dgaussians} and extending single-view inputs to sparse multi-view configurations, facilitating comprehensive 3D object information.

% 已用GPT润色
To address the question of whether the current prompt-based or image-based 3D generation methods are adequate to fulfill our requirements, we can delve further into the necessities of 3D generation and categorize the issue into two distinct subproblems:
% 已用GPT润色
\textbf{(1) Is 3D Generation Controllable?}
In text-to-3D approaches, the prompt typically offers a basic description, requiring users to repeatedly input prompts to achieve the desired 3D output. 
Conversely, image-based methods necessitate acquiring the specific target image that meets the requirements before generating the desired 3D content. 
Therefore, integrating controllability into the 3D generation processes is crucial for ensuring user agency and customization.
% 已用GPT润色
\textbf{(2) Is 3D Generation Efficient?}
The optimization processes involved in text-to-3D and image-to-3D techniques are laborious and time-intensive, often demanding up to an hour to create a single 3D object based on input prompts or images. 
Such extensive computational requirements pose a significant barrier, rendering the production of 3D content unfeasible for many users. 
Consequently, addressing efficiency within the realm of 3D generation stands as a critical challenge to overcome.

% 已用GPT润色
To address the challenges identified, \textbf{this paper aims to develop an efficient and controllable 3D generation method}. 
An existing study named MVControl \cite{li2024controllable} endeavors to tackle this issue by extending ControlNet \cite{zhang2023adding} to a multi-view diffusion model, MVDream \cite{shi2023mvdream}. 
The MVControl system produces four multi-view images, which are then fed into a multi-view Gaussian reconstruction model, LGM \cite{tang2024lgm}, to derive coarse 3D Gaussian representations. 
Subsequently, these coarse Gaussians undergo SDS optimization guided by a 2D diffusion model to refine the 3D Gaussian outputs. 
Despite demonstrating promising outcomes in 3D content generation, MVControl exhibits several limitations:
\textbf{(1) Misalignment between 2D and 3D Representations}: 
In MVControl, the multi-view images generated by the 2D diffusion model are converted to 3D representations using the LGM reconstruction model. 
However, the direct integration of these distinct models may lead to discrepancies between 2D and 3D representations, as the reconstruction model might struggle to generalize across the generated images.
\textbf{(2) Complex Multi-Stage Procedures Increase Time Consumption}: 
MVControl incorporates a two-stage approach: the initial stage involves the amalgamation of 2D diffusion and 3D reconstruction models, while the subsequent stage encompasses the SDS-based optimization process. 
These intricate multi-stage procedures contribute to a cumbersome and time-intensive generation process.

% 已用GPT润色
These identified challenges prompt the following solutions: 
(1) Resolving the misalignment between 2D and 3D through \textbf{an end-to-end aligned model}; 
(2) Streamlining complex procedures with \textbf{a fast feed-forward model}.
This paper introduces \textbf{ControLRM}, a feed-forward model designed for controllable 3D generation founded on the Large Reconstruction Model (LRM). 
The architecture consists of: 
(1) A 2D condition generator with transformer or diffusion backbone that accept text and 2D visual conditions as input; 
(2) A 2D condition encoder that extract 2D latent features from the output feature of the 2D condition generator; 
(3) A triplane decoder transformer that interacts with the 2D features via cross-attention and generate a triplane-NeRF representation.
Training directly with conditional inputs and ground truth multi-view images from scratch is computationally demanding and challenging. 
Therefore, we propose a joint training framework leveraging the strong priors of a pre-trained LRM model trained on extensive datasets.
In the condition training stage, the condition 2D generator and the cross-attention layer are activated, while the parameters in the triplane decoder remain fixed. 
In the image training phase, both the image encoder and the triplane decoder are activated to ensure the alignment between 2D latents and 3D triplane transformer. 
Rather than utilizing the entire Objaverse \cite{deitke2023objaverse} and MVImgNet \cite{yu2023mvimgnet} datasets like LRM \cite{hong2023lrm}, we opt for a smaller dataset, G-Objaverse \cite{qiu2023richdreamer}, to train our ControLRM.
To ensure unbiased evaluation, we curate evaluation samples from three distinct datasets (G-OBJ, GSO, ABO) rather relying on manual generation.
The quantitative and qualitative results on 3D controllability evaluation and generation quality comparison demonstrate the superiority of our method.

% 已用GPT润色
In summary, our main contributions are as follows:
\begin{itemize}
\item We present ControLRM, a novel framework tailored for controllable 3D generation based on single-view 2D condition and text input.
The model undergoes evaluation across four distinct condition types (edge, depth, normal, scribble), showcasing its robust generalization and diverse controllability features.
\item We introduce an end-to-end feed-forward network architecture for controllable 3D generation. The end-to-end paradigm serves as a natural bridge between 2D latents and 3D triplanes, while the feed-forward network design guarantees rapid inference when compared to existing optimization-based approaches.
\item We present an effective joint training scheme for training the controllable 3D generation model. 
This approach leverages the significant 3D reconstruction capabilities within pretrained LRM to enhance our controllable 3D generation task.
\item Through comprehensive experiments conducted on G-OBJ, GSO, and ABO datasets, we demonstrate that our ControLRM significantly surpasses the performance of current state-of-the-art (SOTA) methods in 3D controllability, generation quality, and inference speed (as shown in Fig. \ref{fig:introduction}). 
\end{itemize}

\section{Related Work}
\label{sec-related-work}

\subsection{Optimization-based 3D Generation}

% 已用GPT润色
Building on the accomplishments of text-to-image diffusion models \cite{rombach2022high,liu2023zero}, optimization-based approaches present a practical alternative by circumventing the necessity for extensive text-3D datasets.
DreamFusion \cite{poole2022dreamfusion} is a seminal work that introduced the SDS loss to optimize a neural field using diffusion priors for 3D asset generation. 
Additionally, Score Jacobian Chaining \cite{wang2022scorejacobianchaininglifting} is a study that elevates pretrained 2D diffusion models for 3D creation, utilizing the chain rule and the gradients learned from a diffusion model to backpropagate scores through the Jacobian of a differentiable renderer. 
However, these optimization-based techniques commonly encounter a shared challenge known as the Janus problem. 
MVDream \cite{shi2023mvdream} tackles this issue by refining a multi-view diffusion model, which replaces self-attention with multi-view attention in Unet to produce consistent multi-view images. 
Introducing the concept of 3D Gaussian splatting \cite{kerbl3Dgaussians}, DreamGaussian \cite{tang2023dreamgaussian} optimizes 3D Gaussians using the SDS loss. 
Nonetheless, it grapples with the Janus problem stemming from the uncertainties of 2D SDS supervision and rapid convergence. 
Addressing this, GSGEN \cite{chen2024text} and GaussianDreamer \cite{yi2024gaussiandreamer} incorporate a coarse 3D prior to generate more cohesive geometries. 
Furthermore, GSGEN proposes the use of the 3D SDS loss from Point-E \cite{nichol2022pointegenerating3dpoint} for joint optimization in the geometry phase. 
Despite SDS's benefits in terms of data requirements, it necessitates optimization for each new 3D object and demands hours to reach convergence.

\subsection{Feed-forward 3D Generation}

% 注:之前的内容太多了,只能压缩一下了.
% 已用GPT润色
The extensive 3D datasets \cite{deitke2023objaverse,yu2023mvimgnet} have unlocked new possibilities for training feed-forward models to generate 3D assets directly from text, single- or multi-view images.
(1) \textbf{3D generation from single-view}:
LRM \cite{hong2023lrm} first scales up the triplane transformer on a large dataset to predict a triplane neural radiance field (NeRF) from single-view images, showing high generalization ability.
TripoSR \cite{tochilkin2024triposr} integrates significant improvements in data processing, model design, and training techniques, enhancing the efficiency and effectiveness.
% TGS \cite{zou2024triplane} utilizes a transformer-based point decoder and triplane decoder to reconstruct 3D objects using a hybrid Triplane-Gaussian intermediate representation.
% In addition to transformer-based architectures, Stable Video 3D (SV3D) \cite{voleti2024sv3dnovelmultiviewsynthesis} introduces a latent video diffusion model for high-resolution, image-to-multi-view generation of orbital videos around a 3D object, bridging the gap between 3D generation methods and video diffusion models.
% 3DTopia \cite{hong20243dtopia} combines the strengths of feed-forward networks and optimization-based methods through a two-stage text-to-3D generation system.
(2) \textbf{3D generation from multi-view}:
Methods based on multi-view are extensions designed to enhance the generation quality of single-view methods. Typically, multi-view images of an object are initially synthesized from a single image using a multi-view diffusion model \cite{shi2023mvdream}.
Similar to single-view approaches, these methods can be broadly categorized as either diffusion-based or transformer-based architectures.
Examples of diffusion-based architectures include SyncDreamer \cite{liu2023syncdreamer} and Wonder3D \cite{long2024wonder3d}.
SyncDreamer necessitates dense views for 3D reconstruction, while Wonder3D employs a multiview cross-domain attention mechanism to process relatively sparse views.
Transformer-based architectures like Instant3D \cite{li2023instant3dfasttextto3dsparseview} encodes multi-view images by a image encoder and concatenate the encoded results into a set of tokens for the image-to-triplane decoder.
% InstantMesh \cite{xu2024instantmeshefficient3dmesh} introduces a mesh-based representation and additional geometric supervisions, significantly enhancing training efficiency and reconstruction quality.
Additionally, LGM \cite{tang2024lgm}, GRM\cite{xu2024grmlargegaussianreconstruction} and GS-LRM \cite{zhang2024gs} enhance the generation quality using high-resolution features and increasing the number of surrounding views.
(3) \textbf{3D generation from text}: 
Point-E \cite{nichol2022pointegenerating3dpoint} and Shap-E \cite{jun2023shapegeneratingconditional3d} utilize complex prompts to generate point clouds and neural radiance fields respectively. 
Representing 3D data as volumes, 3DTopia \cite{hong20243dtopia} and VolumeDiffusion \cite{tang2023volumediffusion} train diffusion models by fitting volumetric modules. 
ATT3D \cite{lorraine2023att3damortizedtextto3dobject} employs a feed-forward transformer to generate the 3D contents and train the model with amortized training via pretrained diffusion model.
Latte3D \cite{xie2024latte3d} extends the amortization architecture of ATT3D, significantly improving the efficiency and generation quality.

\begin{figure*}[t]
\centering
\includegraphics[width=\linewidth]{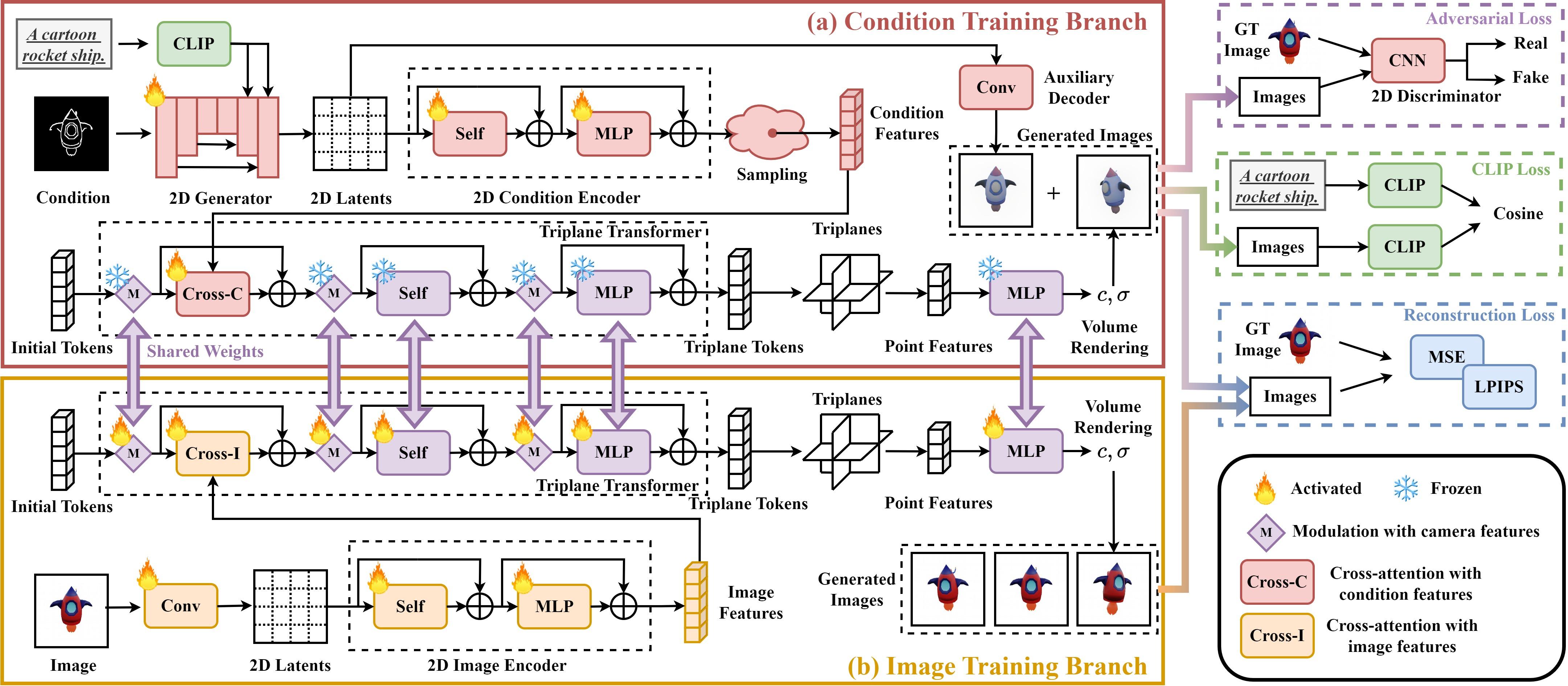}
\vspace{-0.6cm}
\caption{The overall framework of \textbf{ControLRM}, a feed-forward controllable 3D generation model. }
\vspace{-0.4cm}
\label{fig:method}
\end{figure*}

\subsection{Controllable 3D Generation}

% 已用GPT润色
Despite the rapid advancements in 3D generation techniques discussed earlier, achieving controllability in 3D generation remains a significant challenge. 
The current state-of-the-art controllable 3D generation method is MVControl \cite{li2024controllable}. 
This method incorporates a trainable control network that interacts with the base multi-view diffusion model to facilitate controllable multi-view image generation.
In the coarse stage, the MVControl model produces four-view images, which are subsequently input into the 3D reconstruction model LGM \cite{tang2024lgm}. 
The generated coarse Gaussians are then utilized to initialize the SDS-based training in the refinement stage.
% 已用GPT润色
However, there are still some limitations in MVControl: 
(1) The direct integration of distinct models may lead to discrepancies between 2D and 3D representations, as the reconstruction model may not generalize well on the generated multi-view images. 
(2) The complex procedures for generating a single 3D content may increase the time consumption.
In response to these limitations, we propose ControLRM, an end-to-end feed-forward controllable 3D generation model which also has fast inference speed.

\section{Method}
\label{sec-method}

In this section, we present the ControLRM framework as depicted in Fig. \ref{fig:method}. 
We commence by outlining the fundamentals of LRM in Sec. \ref{method:prelinimary_lrm}. 
Next, we delve into a comprehensive examination of the LRM framework from the perspective of the Variational Auto-encoder (VAE) in Sec. \ref{method:lrm_vae}. 
Building on the insights from Sec. \ref{method:lrm_vae}, we elucidate the process of enhancing the LRM to our proposed ControLRM in Sec. \ref{method:upgrade_lrm_to_controlrm}. 
Subsequently, we elaborate on the components of each module within ControLRM and expound on the training objectives in Sec. \ref{method:controlrm}.

% 大纲：
% 1、Preliminary of LRM: 
% 主要介绍LRM的基本组件，用公式描述出来；
% image encoder；
% image-to-triplane decoder；
% 2、Understanding LRM in a Perscpective of VAE
% 这一部分可以基本采用周博之前写好的那些公式，格式稍微调整一下，另外加一些对公式内容的解释；
% 3、Upgrading LRM-VAE to ControLRM
% 这一部分理论后续还要跟周博对一下，需要把joint training加进去似然函数的推导；
% 主要目的还是找到一个可以证明使用基本的图像重建loss训练controlrm branch和lrm branch是能通过elbo保证收敛的；
% 3、ControLRM的各个模块具体介绍
% condition encoder（可以加两个图，ControLRM-T的架构以及ControLRM-D的架构）
% condition-to-triplane decoder
% 具体的重建loss：MSE、SSIM、LPIPS监督损失；CLIP loss；Adversarial loss；
% 4、Training-free Multi-condition Generation

\subsection{Preliminary of LRM}
\label{method:prelinimary_lrm}

Large Reconstruction Model (LRM) is an advanced method that efficiently generates a 3D object from a single 2D image input. 
The LRM primarily consists of the following components:

% Large Reconstruction Model (LRM) is a cutting-edge method engaging in generating 3D object from a single 2D image input with remarkable efficiency.
% Specifically, LRM mainly contains the following components: 

\noindent\textbf{Image Encoder}:
Given an RGB image as input, we utilize a pre-trained visual transformer (ViT) \cite{dosovitskiy2020image} to encode the image into patch-wise feature tokens denoted by $\{h_i | h_i \in \mathbb{R}^{D_e}\}_i^{N_{p}}$, where $i$ represents the index of the image patch, $N_p$ is the total number of image patches, and $D_e$ signifies the dimension of the feature tokens. Specifically, the pre-trained self-supervised model DINO (Caron et al., 2021) is used. 
The ViT incorporates a predefined \texttt{[CLS]} token $h_{\text{cls}} \in \mathbb{R}^{D_e}$, which is then concatenated with the feature sequence $\{h_i\}_{i=1}^{N_{p}}$ to form the output.

% Given an RGB image as input, a pre-trained visual transformer (ViT) \cite{dosovitskiy2020image} is used to encode the image to patch-wise feature tokens 
% $\{h_i|h_i \in \mathbb{R}^{D_e} \}_i^{N_{p}}$
% , where $i$ is the index of image patch, $N_p$ is the total number of image patches, and $D_e$ is the dimension of feature tokens.
% Specifically, the pre-trained self-supervised model DINO \cite{caron2021emerging} is utilized.
% The ViT predefined \texttt{[CLS]} token $h_{\text{cls}} \in \mathbb{R}^{D_e}$ is concatenated with the feature sequence $\{h_i\}_{i=1}^{N_{p}}$ to construct the output.

\noindent\textbf{Camera Features}:
The camera feature $c \in \mathbb{R}^{20}$ is comprised of the flattened vectors of camera extrinsic and intrinsic parameters. 
The 4-by-4 extrinsic matrix $E$ is flattened to a 16-dimensional vector $E_{1 \times 16}$. 
The intrinsic parameters, including the camera focal length and principal points, are combined as a 4-dimensional vector: $[\text{foc}_x, \text{foc}_y, \text{pp}_x, \text{pp}_y]$. 
To embed the camera feature, a multi-layer perceptron (MLP) is employed to transform the camera feature $c$ into a 1024-dimensional camera embedding $\tilde{c}$.
\begin{small}
\begin{equation}
    \tilde{c} = \text{MLP}_{\text{cam}} ( c ) = \text{MLP}_{\text{cam}} ([E_{1\times16}, \text{foc}_x, \text{foc}_y, \text{pp}_x, \text{pp}_y])
    \label{eq1}
\end{equation}
\end{small}

\noindent\textbf{Modulation with Camera Features}:
The camera modulation incorporates an adaptive layer normalization (adaLN) \cite{peebles2023scalable} to adjust image features using denoising iterations and class designations. 
When provided with the camera feature $\tilde{c}$ as input, a multi-layer perceptron (MLP) predicts the scaling factor $\gamma$ and the shifting factor $\beta$:
\begin{small}
    \begin{equation}
        \gamma, \beta = \text{MLP}_{\text{mod}} (\tilde{c})
        \label{eq2}
    \end{equation}
\end{small}

Subsequently, the modulation function will process the sequence of vectors in the transformer $\{f_j\}$ as follows:
% Then, the modulation function will handle the sequence of vectors in transformer $\{f_j\}$ as follows:
\begin{small}
    \begin{equation}
        \text{ModLN}(f_j) = \text{LN}(f_j) \cdot (1 + \gamma) + \beta
        \label{eq3}
    \end{equation}
\end{small}
where $\text{LN}$ is the layer Normalization \cite{lei2016layer}.

\noindent\textbf{Transformer Layers}: 
Each transformer layer consists of a cross-attention sub-layer, a self-attention sub-layer, and a multi-layer perceptron sub-layer (MLP), where the input tokens for each sub-layer are modulated by the camera features. 
The feature sequence $f^{\text{in}}$, serving as the input to the transformer layers, can also be viewed as triplane hidden features. 
As illustrated in Fig. \ref{fig:method} (b), the cross-attention module uses the feature sequence $f_{\text{in}}$ as the query and the image features $\{h_{\text{cls}}, h_i\}_{i=1}^{N_{p}}$ as the key/value pairs.
\begin{small}
    \begin{equation}
        f_j^{\text{cross-i}} = \text{Cross-I} ( \text{ModLN}(f_j^{\text{in}}); \{h_{\text{cls}}, h_i\}_{i=1}^{N_{p}}) + f_j^{\text{in}}
        \label{eq4}
    \end{equation}
\end{small}
where $\text{Cross-I}$ represents the cross-attention between the image features and the triplane features.

Subsequent to the original transformer \cite{vaswani2017attention}, the self-attention sub-layer denoted as $\text{Self}(\cdot)$ and the multi-layer perceptron sub-layer labeled as $\text{MLP}(\cdot)$ handle the input feature sequence in the ensuing manner:
\begin{small}
    \begin{equation}
        f_j^{\text{self}} = \text{Self} (\text{ModLN}(f_j^{\text{cross-i}}); \text{ModLN}(f_{j^{\prime}}^{\text{cross-i}})) + f_j^{\text{cross-i}}
        \label{eq5}
    \end{equation}
\end{small}
\begin{small}
    \begin{equation}
        f_j^{\text{out}} = \text{MLP} (\text{ModLN} (f_j^{\text{self}})) + f_j^{\text{self}}
        \label{eq6}
    \end{equation}
\end{small}
where $f_j^{\text{out}}$ represents the triplane feature output. 
This final output undergoes upsampling via a trainable de-convolution layer and is subsequently reshaped into the final triplane representation $TP \in \mathbb{R}^{3 \times 64 \times 64 \times D_t}$, where $D_t$ signifies the dimension of the triplane.

\noindent\textbf{Triplane NeRF}:
The triplane $\text{TP}$ comprises three axis-aligned feature planes: $\text{TP}_{xy}/\text{TP}_{yz}/\text{TP}_{xz} \in \mathbb{R}^{64 \times 64 \times D_t}$. 
Given any 3D point $p=[p_x,p_y,p_z]^T$ within the NeRF object bounding box $[-1, 1]^3$, the point's feature can be extracted from the triplane $\text{TP}$ using bilinear sampling.
\begin{small}
    \begin{equation}
        \text{TP}_p = \text{Concat} (\text{TP}_{xy}[p_x,p_y], \text{TP}_{yz}[p_y,p_z], \text{TP}_{xz}[p_x,p_z])
        \label{eq7}
    \end{equation}
\end{small}
where $\text{Concat}(\cdot)$ represents the concatenation function, and $\text{TP}_p \in \mathbb{R}^{3 \cdot D_t}$ denotes the sampled feature corresponding to point $p$.

\noindent\textbf{Training Objectives}:
During training, $V$ views are randomly selected from the dataset. 
One view is chosen as the reference view and passed to the LRM, while the other $V-1$ views serve as auxiliary training views. 
Let the rendered views of the LRM be denoted as $\hat{x}$, and the ground truth views as $x^{\text{GT}}$. 
Particularly, for each input image $x$, we aim to minimize:
\begin{small}
    \begin{equation}
        L_{\text{recon}} (x) = \frac{1}{V} \sum_{v=0}^{V} (L_{\text{MSE}} (\hat{x}_v, x^{\text{GT}}_v) + \lambda L_{\text{LPIPS}} (\hat{x}_v, x^{\text{GT}}_v))
        \label{eq8}
    \end{equation}
\end{small}
where $L_{\text{MSE}}$ represents the normalized pixelwise L2 loss, $L_{\text{LPIPS}}$ denotes the perceptual image similarity loss \cite{zhang2018unreasonable}, and $\lambda$ is a customizable weight used to balance these losses.

% 周博写的版本
% The Large Reconstruction Model(LRM) is a cutting-edge system that engage in generating 3D object from a single 2D image with remarkable speed and efficiency.
% The first innovative approach was proposed by \cite{hong2023lrm}, which was built on a transformer-based architecture.
% Specifically, LRM mainly contains the following components: Image Encoder, Image-to-Triplane Decoder and Neural Radiance Field (NeRF).
% Image Encoder is a pretrained visual transformer \cite{dosovitskiy2020image,caron2021emerging} which take the image as input and output the image feature tokens to capture the image's structural and textural details. 
% The encoded image feature tokens are then projected onto a 3D triplane representation \cite{chan2022efficient,gao2022get3d} through the Image-to-Triplane Decoder which applies cross-attention to align image features with the triplane and self-attention to refine the internal structure within the triplane.
% Moreover, camera parameters are formulated as features within decoding process which account for viewing perspective to modulate the triplane features. 
% With the refined triplane representation, the color and density of each point in 3D space is predicted within NeRF.
% The final step involves rendering images at arbitrary views by utilizing the output of NeRF.
% The entire pipeline is trained holistically end-to-end on a vast dataset, using a simple reconstruction losses \cite{zhang2018unreasonable} to minimize the differences between rendered and original input views.

\subsection{Understanding LRM in a Perspective of VAE}
\label{method:lrm_vae}

From the perspective of Variational Autoencoder (VAE) \cite{kingma2022autoencoding}, the LRM can be viewed as an intricate architecture that encompasses certain fundamental principles akin to VAEs.

Similar to the encoder in a VAE, the image encoder of LRM processes an input image, transforming it into a series of feature tokens.
These tokens serve as the encoded latent representation of the input image, mirroring the latent space in a VAE.
The decoding component of LRM functions analogously to the decoder in a VAE by reconstructing images from the latent space.
Specifically, LRM maps the latent trilinear representation to a 3D object within NeRF and subsequently generates images with new perspectives, akin to the generation or decoding process within a VAE framework.
LRM employs a reconstruction loss to reduce the dissimilarity between the input image and the rendered images altered based on camera parameters. 
In the subsequent section, we will offer a theoretical overview of LRM, including a form of Evidence Lower Bound (ELBO).

Given the 3D representation $\bold{x}_{3d}$, a set of projected 2D images $\{x_i\}_{i=1}^{N_V}$ with corresponding camera parameters $\{T_i\}_{i=1}^{N_V}$, where $N_V$ denotes the number of viewpoints. 
It is assumed that the ground-truth distribution of the 3D representation is represented by the density $p(\bold{x}_{3d})$. 
In LRM, this 3D representation is characterized by a triplane Neural Radiance Field (NeRF). 
Under this assumption, one can write:
\begin{small}
\begin{equation}
p(\bold{x}_{3d}) = \int_{z} p(\bold{x}_{3d},z)dz=\int_{z}p(\bold{x}_{3d}|z)p(z)dz
\label{eq9}
\end{equation}
\end{small}
$z$ represents the latent variable associated with $\bold{x}_{3d}$, following a simple distribution $p(z)$ referred to as the prior distribution. 
The primary objective of the VAE is to acquire a robust approximation of $p(\bold{x}_{3d}|z)$ based on the provided data. 
This approximated distribution is denoted by $p_{\theta}(\bold{x}_{3d}|z)$, where $\theta$ symbolizes the learnable parameters. 
Subsequently, we can compute the log likelihood $\log{p_\theta(\mathbf{x}_{3d})}$ in the following manner:
\begin{equation}
\resizebox{0.7\linewidth}{!}{$
\begin{aligned}
&\log{p_{\theta}(\bold{x}_{3d})}\\=&\log{\int_{T}}p_{\theta}(\bold{x}_{3d}|T)p(T)dT
\geq \int_{T}\log{p_{\theta}(\bold{x}_{3d}|T)}p(T)dT\\
\approx&\frac{1}{N_V}\sum\limits_{i=1}^{N_V}\log{p_{\theta}(\bold{x}_{3d}|T_{i})}
=\frac{1}{N_V}\sum\limits_{i=1}^{N_V}\log{\int_{z}}p_{\theta}(\bold{x}_{3d},z|T_{i})dz\\
=&\frac{1}{N_V}\sum\limits_{i=1}^{N_V}\log\int_{z}\frac{p_{\theta}(\bold{x}_{3d},z|T_{i})q_{\varphi}(z|x_{i},T_{i})}{q_{\varphi}(z|x_{i},T_{i})}dz\\
\geq&\frac{1}{N_V}\sum\limits_{i=1}^{N_V}\mathbb{E}_{q_{\varphi}}\log{\frac{p_{\theta}(\bold{x}_{3d},z|T_{i})}{q_{\varphi}(z|x_{i},T_{i})}}\\
\end{aligned}
$}
\label{eq10}
\end{equation}
where $p_\theta(\bold{x}_{3d}|T_i)$ indicates that the 3D representation $\bold{x}_{3d}$ is conditioned on the camera parameters $T_i$ corresponding to viewpoint $i$. 
Given that our $\bold{x}_{3d}$ embodies a triplane NeRF, when conditioned on $T_i$, it serves as a representation of the rendered image from viewpoint $i$. 
The final row in Eq. \ref{eq10} denotes the Evidence Lower Bound (ELBO). 
By isolating the inner term of ELBO at viewpoint $i$, we obtain:
\begin{equation}
\resizebox{0.8\linewidth}{!}{$
\begin{aligned}
&\mathbb{E}_{q_{\varphi}}\log{\frac{p_{\theta}(\bold{x}_{3d},z|T_{i})}{q_{\varphi}(z|x_{i},T_{i})}}\\
=&\mathbb{E}_{q_{\varphi}}\log\frac{p_{\theta}(\bold{x}_{3d}|z,T_{i})p_{\theta}(z)}{q_{\varphi}(z|x_{i},T_{i})}\\
=&\mathbb{E}_{q_{\varphi}}\log{p_{\theta}(\bold{x}_{3d}|z,T_{i})}
 -\text{KL}(q_{\varphi}(z|x_{i},T_{i})||p_{\theta}(z))\\
=&\mathbb{E}_{q_{\varphi}}\log{\int_{T}p_{\theta}(\bold{x}_{3d}|z,T_{i},T)}p(T)dT
-\text{KL}(q_{\varphi}(z|x_{i},T_{i})||p_{\theta}(z))]\\
\geq&\mathbb{E}_{q_{\varphi}}\int_{T}\log{p_{\theta}(\bold{x}_{3d}|z,T_{i},T)}p(T)dT
-\text{KL}(q_{\varphi}(z|x_{i},T_{i})||p_{\theta}(z))\\
\approx&\frac{1}{M}\sum\limits_{j=1}^{M}\mathbb{E}_{q_{\varphi}}\log{p_{\theta}(\bold{x}_{3d}}|z,T_{i},T_{j})-\text{KL}(q_{\varphi}(z|x_{i},T_{i})||p_{\theta}(z))
\end{aligned}
$}
\label{eq11}
\end{equation}

Note that the extrinsic matrix of the input reference view is normalized to an identity matrix, while the extrinsic matrices of the other views are adjusted to the relative transformation matrix with respect to the normalized reference view.
The intrinsic parameters remain constant across all views. 
Consequently, the input camera parameter $T_i$ is consistent and fixed within the LRM, thereby allowing for its exclusion from the formulas:
\begin{small}
\begin{equation}
\begin{aligned}
&\mathbb{E}_{q_{\varphi}}\log{\frac{p_{\theta}(\bold{x}_{3d},z|T_{i})}{q_{\varphi}(z|x_{i},T_{i})}}\\
\geq &\frac{1}{M}\sum\limits_{j=1}^{M}\mathbb{E}_{q_{\varphi}}\log{p_{\theta}(\bold{x}_{3d}}|z,T_{j})-\text{KL}(q_{\varphi}(z|x_{i})||p_{\theta}(z))
\label{eq12}
\end{aligned}
\end{equation}
\end{small}
where $p_{\theta}(x_{3d}|z,T_j)$ represents the triplane decoder (depicted as purple modules in Fig. \ref{fig:method}), while $q_{\phi}(z|x_i)$ denotes the image encoder (illustrated as orange modules in Fig. \ref{fig:method}).

\subsection{Upgrading LRM to ControLRM}
\label{method:upgrade_lrm_to_controlrm}

Eq.  \ref{eq10}, \ref{eq11}, and \ref{eq12} elaborate on the extension of LRM, interpreting it as a specialized variant of the Variational Autoencoder (VAE).
By analogy, these expressions can be further expanded to cater to the objective of controllable 3D generation.
Consider $e_i$ as indicative of the input 2D visual condition on view $i$ and the  associated textual prompt concerning the 3D object, the ELBO can be formulated as:
\begin{equation}
\resizebox{0.8\linewidth}{!}{$
\begin{aligned}
&\log{p_{\theta}(\bold{x}_{3d})}\\=&\log{\int_{T}}p_{\theta}(\bold{x}_{3d}|T)p(T)dT
\geq \int_{T}\log{p_{\theta}(\bold{x}_{3d}|T)}p(T)dT\\
\approx&\frac{1}{N_V}\sum\limits_{i=1}^{N_V}\log{p_{\theta}(\bold{x}_{3d}|T_{i})}
=\frac{1}{N_V}\sum\limits_{i=1}^{N_V}\log{\int_{z}}p_{\theta}(\bold{x}_{3d},z|T_{i})dz\\
=&\frac{1}{N_V}\sum\limits_{i=1}^{N_V}\log\int_{z}\frac{p_{\theta}(\bold{x}_{3d},z|T_{i})q_{\varphi'}(z|e_{i},T_{i})}{q_{\varphi'}(z|e_{i},T_{i})}dz\\
=&\frac{1}{N_V}\sum\limits_{i=1}^{N_V}\log{\mathbb{E}_{q_{\varphi'}}}[\frac{p_{\theta}(\bold{x}_{3d},z|T_{i})}{q_{\varphi'}(z|e_{i},T_{i})}]\geq\frac{1}{N_V}\sum\limits_{i=1}^{N_V}\mathbb{E}_{q_{\varphi'}}\log{\frac{p_{\theta}(\bold{x}_{3d},z|T_{i})}{q_{\varphi'}(z|e_{i},T_{i})}}\\
\end{aligned}
$}
\label{eq13}
\end{equation}

By isolating the inner term of ELBO at viewpoint $i$, we can get: 
\begin{equation}
\resizebox{0.9\linewidth}{!}{$
\begin{aligned}
&\mathbb{E}_{q_{\varphi'}(z|e_{i},T_{i})}\log{\frac{p_{\theta}(\bold{x}_{3d},z|T_{i})}{q_{\varphi'}(z|e_{i},T_{i})}}\\
=&\mathbb{E}_{q_{\varphi'}(z|c_{i},T_{i})}\log\frac{p_{\theta}(\bold{x}_{3d}|z,T_{i})p_{\theta}(z)}{q_{\varphi'}(z|e_{i},T_{i})}\\
\geq&\frac{1}{M}\sum\limits_{j=1}^{M}\mathbb{E}_{q_{\varphi'}(z|e_{i},T_{i})}\log{p_{\theta}(\bold{x}_{3d}}|z,T_{i},T_{j})-\text{KL}(q_{\varphi'}(z|e_{i},T_{i})||p_{\theta}(z))
\end{aligned}
$}
\label{eq14}
\end{equation}

Due to the normalization operation towards reference viewe in the extrinsic matrix, $T_i$ is a fixed identity matrix, which can be further simplified in Eq. \ref{eq14}.
\begin{equation}
\resizebox{0.9\linewidth}{!}{$
\begin{aligned}
&\mathbb{E}_{q_{\varphi'}(z|e_{i},T_{i})}\log{\frac{p_{\theta}(\bold{x}_{3d},z|T_{i})}{q_{\varphi'}(z|e_{i},T_{i})}}\\
\geq&\frac{1}{M}\sum\limits_{j=1}^{M}\mathbb{E}_{q_{\varphi'}(z|e_{i})}\log{p_{\theta}(\bold{x}_{3d}}|z,T_{j})-\text{KL}(q_{\varphi'}(z|e_{i})||p_{\theta}(z))
\end{aligned}
$}
\label{eq15}
\end{equation}
where $p_{\theta}(x_{3d}|z,T_j)$ represents the same triplane decoder as Eq. \ref{eq12} (depicted as purple modules in Fig. \ref{fig:method}), while $q_{\varphi'}(z|e_i)$ denotes the condition encoder part (illustrated as red modules in Fig. \ref{fig:method}).

Eq. \ref{eq15} represents the ELBO of our ControLRM.
However, the optimization of Eq. \ref{eq15} might be much more difficult than the optimization of Eq. \ref{eq12} in LRM, given the relaxation of input from detailed images to coarse conditions (visual condition maps and text descriptions).
Typically, achieving convergence of ControLRM necessitates an even larger scale of data compared to what was utilized in training LRM (Objaverse \cite{deitke2023objaverse} and MVImgNet \cite{yu2023mvimgnet}).
Consequently, direct optimization of Eq. (\ref{eq15}) is not the optimal solution, considering the computational cost and convergence issues encountered during training.

To address this issue, we have to explore an alternative training approach for our ControLRM model. 
Remarkably, it is observed that the triplane decoder denoted by $p_{\theta}(x_{3d}|z,T_j)$ is common to both Eq. \ref{eq12} and Eq. \ref{eq15}. 
This implies that leveraging the convergence of the triplane decoder $p_{\theta}(x_{3d}|z,T_j)$ and the image encoder $q_{\phi}(z|x_i)$ under the guidance of Eq. (\ref{eq12}) can enhance the training process in Eq. \ref{eq15}. 
If $p_{\theta}(x_{3d}|z,T_j)$ is kept constant in Eq. \ref{eq15}, the focus shifts to maximizing the remaining term $-\text{KL}(q_{\varphi'}(z|e_{i})||p_{\theta}(z))$, aligning the condition encoder $q_{\varphi'}(z|e_{i})$ with the latent space $z$.
Consequently, the need for a vast amount of paired data (input condition and 3D object) can be significantly reduced, and the convergence can also be enhanced by leveraging the strong prior knowledge embedded in pretrained LRM models.

Following these discussions, we propose a joint training paradigm which comprises two branches: the Image Training Branch and the Condition Training Branch.
The former encompasses a 2D image encoder ($q_{\phi}(z|x_i)$ in Eq. \ref{eq12}) and a 3D triplane decoder ($p_{\theta}(x_{3d}|z,T_j)$ in Eq. \ref{eq12}). 
The latter comprises a 2D condition encoder ($q_{\varphi'}(z|e_{i})$ in Eq. \ref{eq15}) and utilizes the same 3D triplane decoder ($p_{\theta}(x_{3d}|z,T_j)$ in Eq. \ref{eq15}. 
It is noteworthy that the cross-attention layers interacting with $q_{\phi}(z|x_i)$ and $q_{\varphi'}(z|e_{i})$ are denoted as Cross-I and Cross-C, respectively. 
Illustrated in Figure \ref{fig:method}, the Image Training Branch optimizes the ELBO in Eq. \ref{eq12}, aiming to refine the triplane decoder $p_{\theta}(x_{3d}|z,T_j)$ and 2D image encoder $q_{\phi}(z|x_i)$ for optimal performance. 
On the other hand, the Condition Training Branch retains the fixed parameters of the triplane decoder $p_{\theta}(x_{3d}|z,T_j)$ and focuses on optimizing the ELBO in Eq. \ref{eq15}. 
This process naturally aligns the distributions of the latent spaces in Eq. \ref{eq12} and Eq. \ref{eq15} using the shared 3D Triplane Transformer.

\subsection{ControLRM}
\label{method:controlrm}

In this section, we delve into the specific modules of ControLRM. 
The design of the conditional generator was detailed in Fig. \ref{fig:method} in Section \ref{method:controlrm:conditional_generator}. 
Depending on the chosen backbone for the conditional generator, ControLRM manifests in two variants: 
1) ControLRM-T featuring a transformer-based conditional generator (Section \ref{method:controlrm:transformer_generator}); 
2) ControLRM-D integrating a diffusion-based conditional generator (Section \ref{method:controlrm:diffusion_generator}). 
Subsequently, we present the condition-to-triplane transformer decoder in Section \ref{method:controlrm:condition_to_triplane}. 
The training objectives encompassing adversarial loss, clip loss, and rendering loss are expounded upon in Section \ref{method:controlrm:training_objectives}.

% In this section, we introduce the details of modules in ControLRM.
% We introduced the design of the conditional generator in Fig. \ref{fig:method} in Sec. \ref{method:controlrm:conditional_generator}.
% Depending on the backbone adopted in the conditional generator, our ControLRM has 2 variations: 1) \textbf{ControLRM-T} with transformer-based conditional generator (Sec. \ref{method:controlrm:transformer_generator}); 2) \textbf{ControLRM-D} with diffusion-based conditional generator (Sec. \ref{method:controlrm:diffusion_generator}).
% Then, we introduce the condition-to-triplane transformer decoder in Sec. \ref{method:controlrm:condition_to_triplane}.
% The training objectives of adversarial loss, clip loss and rendering loss are introduced in Sec. \ref{method:controlrm:training_objectives}.

\subsubsection{Design of Conditional Generator}
\label{method:controlrm:conditional_generator}

As depicted in Fig. \ref{fig:method}, the conditional generator utilizes the 2D condition and the text embedding of CLIP \cite{radford2021learning} as input to produce the 2D latents required for subsequent procedures. 
A naive design of this generator is a transformer-based backbone with cross-attention mechanism between the feature sequence extracted from condition image and the text feature.
However, this design with only the cross-attention mechanism fails to generate a regular results but yielding meaningless results in the experiments.
A similar issue was observed in \cite{li2024instant3d}, indicating that the main reason for this optimization failure stems from the notable disparity between the 2D renderings and the ground truth images. 
As noted by \cite{arjovsky2017towards}, the optimization gradient becomes unreliable when the generated distribution and the target distribution are disjoint. 
In contrast, the backward gradients to the 2D latents in our model must traverse a series of modules, including the condition encoder, triplane transformer, and NeRF modules. 
This complexity of pathways may significantly impede the optimization process, consequently resulting in unexpected failures. 
A straightforward remedy proposed in \cite{li2024instant3d} involves the incorporation of randomness (e.g., Gaussian noise) into the network architecture. 
By increasing the overlap between the rendered distribution and the target distribution, the gradients during training become more meaningful, promoting convergence. In summary, the key considerations for designing the condition generator in ControLRM are:
1) Incorporation of randomness for improved training outcomes. 2) Emphasis on the efficiency of the generator for fast inference speed.

% As shown in Fig. \ref{fig:method}, the conditional generator takes the 2D condition and the text embedding of CLIP \cite{radford2021learning} as input, outputting the 2D latents for the following procedures.
% A naive design of this generator is a transformer-based backbone with cross-attention mechanism between the feature sequence extracted from condition image and the text feature.
% However, this design with only the cross-attention mechanism fails to generate a regular results but yielding meaningless results in the experiments.
% Similar phenomenon is also witnessed in \cite{li2024instant3d}, and the reason of this optimization failure is mainly caused by the significant discrepancy between the 2D renderings and the ground truth image.
% \cite{arjovsky2017towards} mention that the optimization gradient is unreliable when the generated and the target distribution are disjoint.
% Whereas the backwarded gradients to the 2D latents in our model  has to pass through a series of modules including: condition encoder, triplane transformer, and NeRF modules.
% This may make the optimization pretty hard to converge, thus leading to an unexpected failure.
% A simple solution \cite{li2024instant3d} to this problem is integrating randomness (i.e. Gaussian noise) into the network.
% The overlap between the rendered and the target distributions can be enlarged, leading to more meaningful gradients during the training.
% In consequence, we summarize the key points in designing the condition generator as follows:
% 1) Integration of randomness.
% 2) Efficiency of the generator.

\begin{figure}[t]
\centering
\includegraphics[width=\linewidth]{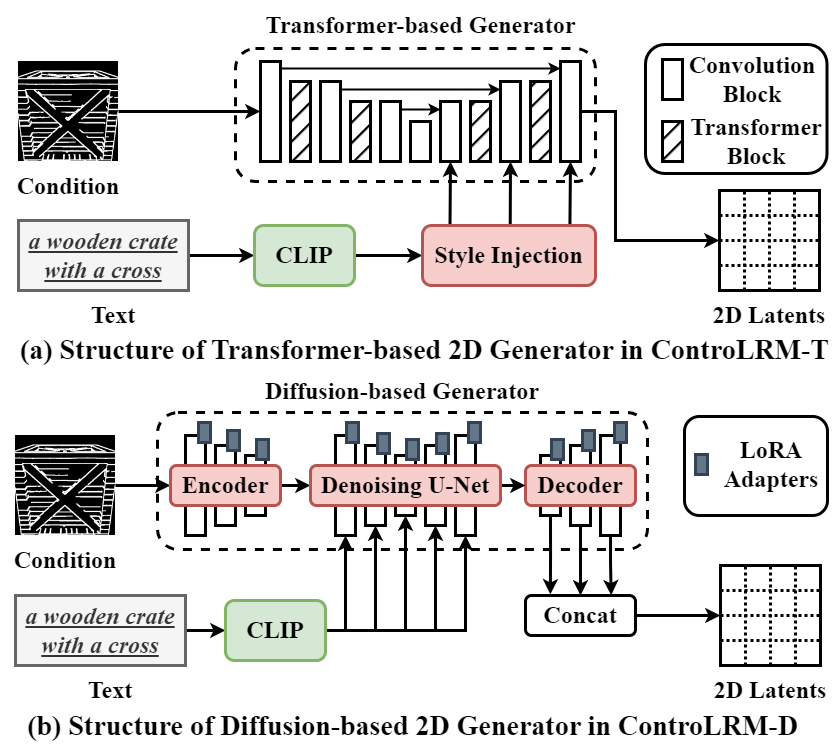}
\vspace{-0.6cm}
\caption{The architecture of the 2D conditional generator in ControLRM. (a) shows the transformer-based generator in \textbf{ControLRM-T}, and (b) shows the diffusion-based generator in \textbf{ControLRM-D}.}
\vspace{-0.4cm}
\label{fig:condition_generator}
\end{figure}

\subsubsection{Transformer-based Conditional Generator}
\label{method:controlrm:transformer_generator}

For \textbf{ControLRM-T} model, we have devised a lightweight transformer-based generator, illustrated in Figure \ref{fig:condition_generator} (a). 
Building upon the preceding discussion, we introduce randomness through a style injection module. 
Drawing inspiration from the original style injection concept in StyleGAN \cite{karras2019style}, where style features and random noise are integrated into the generator via Adaptive Instance Normalization (AdaIN), we adapt this approach by treating the text embedding as the style feature. 
This text embedding is concatenated with random Gaussian noise and passed through a 3-layer MLP within our style injection module. 
The resulting feature vector is then combined with the output of each convolution layer to incorporate the text feature. 
In Figure \ref{fig:condition_generator} (a), the convolution blocks and transformer blocks are stacked together, with residual connections applied to the convolution blocks in a U-Net configuration. 
% For a comprehensive overview of each layer's specifications, please refer to the appendix.

% For \textbf{ControLRM-T}, we design a light-weight transformer-based generator as shown in Fig. \ref{fig:condition_generator} (a).
% Based on the discussion in previous section, we involve the randomness via a style injection module.
% In the original design of style injection in StyleGAN \cite{karras2019style}, the style feature and random noise are injected into the style-based generator via Adaptive Instance Normalization (AdaIN).
% In analogy, we can treat the text embedding as the style feature, and concatenate it with random Gaussian noise.
% Then, the concatenated feature is fed to a 3-layer MLP in our style injection module.
% The output feature vector is further concatenated with the output of each convolution layer to inject the text feature.
% As shown in Fig. \ref{fig:condition_generator} (a), the convolution blocks and transformer blocks are stacked together, and the short-cuts append residual connections on the convolution blocks in a U-Net structure.
% The detailed setting of each layer is provided in the appendix.

\subsubsection{Diffusion-based Conditional Generator}
\label{method:controlrm:diffusion_generator}

For the \textbf{ControLRM-D} model, we have intricately integrated LoRA adapters \cite{hu2021lora} into the original latent diffusion model, incorporating small trainable weights. 
Leveraging the inherent randomness within the diffusion model, and aided by the pre-trained weights obtained from large-scale datasets, we aim to address the discrepancy issue highlighted in Section \ref{method:controlrm:condition_encoder}.
In addressing efficiency concerns, we opt for the fast one-step diffusion model \cite{parmar2024one} as the foundational framework. 
Specifically, we initialize the Diffusion-based generator with the pre-trained weights of SD-Turbo \cite{sauer2024fast}. 
To form the 2D latents for subsequent procedures, we concatenate the outputs of the last three layers of the decoder depicted in Figure \ref{fig:condition_generator} (b). 
% For a more detailed configuration of each layer, please refer to the appendix.

% For \textbf{ControLRM-D}, we tightly integrate LoRA adapters \cite{hu2021lora} into the original latent diffusion model with small trainable weights.
% The diffusion model naturally possesses the integration of randomness, and the pre-trained weights on large-scale dataset can help to remedy the discrepancy problem discussed in Sec. \ref{method:controlrm:condition_encoder}.
% In consideration of the efficiency problem, we choose the fast one-step diffusion model \cite{parmar2024one} as the backbone.
% Specifically, the pre-trained weights of SD-Turbo \cite{sauer2024fast} are used for initialization of the Diffusion-based generator.
% The output of the last 3 layers of the decoder in Fig. \ref{fig:condition_generator} (b) are concatenated together to create the 2D latents for the following procedure.
% Further detailed setting of each layer is provided in the appendix.

\subsubsection{Condition Encoder}
\label{method:controlrm:condition_encoder}

In Figure \ref{fig:method} (a), the 2D latents are firstly interpolated to match the resolution of the input condition image, and then divided into the feature sequence $\{ g_i | g_i \in \mathbb{R}^{D_e} \}_i^{N_p}$. 
Similar to the feature sequence $\{h_i\}_i^{N_p}$ extracted from the input image discussed in Sec. \ref{method:preliminary_lrm}, $D_e$ denotes the feature dimension, while $N_p$ corresponds to the number of patches. 
Within the condition encoder, the feature sequence $\{g_i\}_i^{N_p}$ is passed through a sequence of transformer layers, each comprising a self-attention sub-layer and an MLP sub-layer.
% As shown in Fig. \ref{fig:method} (a), the 2D latents is firstly interpolated to the same resolution as the input condition image, and then split into the feature sequence $\{ g_i | g_i \in \mathbb{R}^{D_e} \}_i^{N_p}$.
% Same as the feature sequence $\{h_i\}_i^{N_p}$ extracted from input image in Sec. \ref{method:prelinimary_lrm}, $D_e$ means the dimension of feature, and $N_p$ represents the number of patches.
% In the condition encoder, the feature sequence $\{g_i\}_i^{N_p}$ is fed to a series of transformer layers where each layer contains a self-attention sub-layer and a MLP sub-layer:
\begin{small}
    \begin{equation}
        g_i^{\text{self}} = \text{Self} ( g_i; g_i  ) + g_i
    \end{equation}
\end{small}
\begin{small}
    \begin{equation}
        g_i^{\text{out}} = \text{MLP} (g_i^{\text{self}}) + g_i^{\text{self}}
    \end{equation}
\end{small}
where $g_i^{\text{out}}$ is the output feature.

To integrate the random sampling process, the output $g_i^{\text{out}}$ of the final transformer layer is fed to another MLP to regress the mean and variance results:
\begin{small}
    \begin{equation}
        \mu_{g_i}, \sigma_{g_i} = \text{MLP} (g_i^{\text{out}})
    \end{equation}
\end{small}
where $\mu_g$ is the mean feature and $\sigma_g$ represents the variance.
Throughout training, the output feature sequence $\{\tilde{g}_i\}_i^{N_p}$ is is stochastically sampled from a Gaussian distribution, where $\tilde{g}_i \sim \mathcal{N}(\mu_{g_i}, \sigma_{g_i}^2)$.

\subsubsection{Auxiliary Decoder}
\label{method:controlrm:auxiliary_decoder}

To boost the performance, we further introduce an auxiliary decoder for the 2D latents to enhance the training process. 
The generated 2D latents from the conditional generator (refer to Sections \ref{method:controlrm:transformer_generator} and \ref{method:controlrm:diffusion_generator}) are passed through a lightweight three-layer convolutional neural network. 
The resulting image $x_{\text{aux}}$ is combined with the 2D renderings to compute the loss function for the generated images. 
The inclusion of the auxiliary decoder offers direct guidance to the 2D generator, aiding in overall network convergence.

% In avoidance of the possible failure discussed in Sec. \ref{method:controlrm:condition_encoder}, we further add an auxiliary decoder to the 2D latents to assist the training process.
% The output 2D latents of the conditional generator (Sec. \ref{method:controlrm:transformer_generator} and \ref{method:controlrm:diffusion_generator}) are fed to a light-weight convolutional neural network with only 3 layers.
% The output image $x_{\text{aux}}$ is added with the 2D renderings as the generated images for loss calculation.
% The existance of the auxiliary decoder can provide a direct guidance to the 2D generator and help the convergence of the whole network.

\subsubsection{Triplane Transformer Decoder}
\label{method:controlrm:condition_to_triplane}

The condition-to-triplane decoder receives the condition feature sequence $\{\tilde{g}_i\}_i^{N_p}$ and the triplane feature sequence $f^{\text{in}}$. 
Analogous to the image-to-triplane decoder discussed in Sec. \ref{method:prelinimary_lrm}, each transformer layer consists of a cross-attention sub-layer, a self-attention sub-layer, and an MLP layer. 
The input tokens for each sub-layer are influenced by the camera features $\tilde{c}$. 
The operation of each transformer layer can be described as follows:
\begin{small}
    \begin{equation}
        f_j^{\text{cross-c}} = \text{Cross-C}(\text{ModLN} (f_j^{\text{in}});\{\tilde{g}_i\}_i^{N_p}) + f_j^{\text{in}}
    \end{equation}
\end{small}
\begin{small}
    \begin{equation}
        f_j^{\text{self}} = \text{Self}(\text{ModLN} (f_j^{\text{cross-c}}); \text{ModLN} (f_j^{\text{cross-c}})) + f_j^{\text{cross-c}}
    \end{equation}
\end{small}
\begin{small}
    \begin{equation}
        f_j^{\text{out}} = \text{MLP} ( \text{ModLN} (f_j^{\text{self}}) ) + f_j^{\text{self}}
    \end{equation}
\end{small}

\subsubsection{Training Objectives}
\label{method:controlrm:training_objectives}

In Fig. \ref{fig:method}, the training objectives consist of three components: adversarial loss, CLIP loss, and rendering loss. 
For each sample, we designate one reference view and randomly select $V-1$ side views. 
Denoting the rendered images of ControLRM as $\hat{x}$ and the ground truth images as $x^{\text{GT}}$, the index of the reference view is designated as $0$.
The resultant image from the auxiliary decoder (refer to Sec. \ref{method:controlrm:auxiliary_decoder}) is denoted as $x_{\text{aux}}$. 
The calculation of the loss can be expressed as follows:

\noindent\textbf{Adversarial Loss:}
To incentivize the alignment of the generated images with the corresponding ground truth domains, we apply an adversarial loss \cite{goodfellow2014generative}. 
In line with the approach advocated by Vision-Aided GAN \cite{kumari2022ensembling}, the discriminator utilizes the CLIP model as its foundation. 
The adversarial loss is defined as follows:
% We use an adversarial loss \cite{goodfellow2014generative} to encourage the generated images to match the corresponding ground truth domains.
% Following the recommendations of Vision-Aided GAN \cite{kumari2022ensembling}, the discriminator use the CLIP model as the backbone.
% The adversarial loss can be defined as:
\begin{small}
    \begin{equation}
    \begin{aligned}
        L_{\text{adv}} =& \frac{1}{V+1} \{ \sum_{v=0}^{V} \mathbb{E} [\log \mathcal{D} (x^{\text{GT}}_v)] + \sum_{v=0}^{V} \mathbb{E} [ \log(1 - \mathcal{D}( \hat{x}_v )) ] + \\
        & \mathbb{E} [\log \mathcal{D} (x^{\text{GT}}_0)] + \mathbb{E} [ \log(1 - \mathcal{D}( x_{\text{aux}} )) \} 
    \end{aligned}
    \end{equation}
\end{small}

\noindent\textbf{CLIP Loss:}
To improve the consistency between the generated images and the text prompt $y_{\text{text}}$, a CLIP loss \cite{radford2021learning} is employed for text-image alignment.
% We use a CLIP \cite{radford2021learning} text-image alignment loss to enhance the consistency between generated images and the text prompt $y_{\text{text}}$.
\begin{small}
    \begin{equation}
    \begin{aligned}
        L_{\text{clip}} =& \frac{1}{V+1} [ \sum_{v=0}^{V} (1 - \cos ( \text{CLIP-I}(\hat{x}_v), \text{CLIP-T}(y_{\text{text}}))) + \\
        & (1 - \cos (\text{CLIP-I} (x_{\text{aux}}), \text{CLIP-T}(y_{\text{text}}))) ]
    \end{aligned}
    \end{equation}
\end{small}
where $\text{CLIP-I}$ is the CLIP image encoder, and $\text{CLIP-T}$ is the CLIP text encoder.

\noindent\textbf{Reconstruction Loss:}
The generated images are compared to the ground truth images to ensure consistency through a reconstruction loss. 
For each input condition image and text prompt, we aim to minimize:
% The generated images and ground truth images are enforced to be consistent via a reconstruction loss.
% For every input condition image and text prompt, we minimize:
\begin{small}
    \begin{equation}
    \begin{aligned}
        L_{\text{recon}} =& \frac{1}{V+1} [ \sum_{v=0}^{V} (L_{\text{MSE}} (\hat{x}_v, x^{\text{GT}}_v) + \lambda \sum_{v=0}^{V} L_{\text{LPIPS}} (\hat{x}_v, x^{\text{GT}}_v)) +\\
        & L_{\text{MSE}} (x_{\text{aux}},x^{\text{GT}}_{0}) + \lambda L_{\text{LPIPS}} (x_{\text{aux}},x^{\text{GT}}_{0}) ]
    \end{aligned}
    \end{equation}
\end{small}
where $L_{\text{MSE}}$ is the normalized pixel-wise L2 loss, $L_{\text{LPIPS}}$ is the perceptual image patch similarity \cite{zhang2018unreasonable}.
$\lambda$ is a customized weight to balance the losses.
In default, $\lambda = 1.0$.

\noindent\textbf{Overall Loss:}
The overall loss is a weighted sum of the aforementioned losses:
\begin{small}
    \begin{equation}
        L_{\text{overall}} = L_{\text{recon}} + \lambda_{\text{adv}} L_{\text{adv}} + \lambda_{\text{clip}} L_{\text{clip}}
    \end{equation}
\end{small}
where $\lambda_{\text{adv}} = 0.5$, $\lambda_{\text{clip}} = 5.0$ in default.

\noindent\textbf{Efficient Training:}
By default, we configure the rendered image resolution to $256 \times 256$. 
However, performing direct computations on the entire $256 \times 256$ renderings using $L_{\text{overall}}$ is likely to lead to GPU memory overflow during training, mainly due to NeRF's significant memory requirements. 
To address this issue, we opt for a straightforward yet efficient approach that trades space for time. 
Firstly, we partition the original $256 \times 256$ images into smaller local patches with a resolution of $128 \times 128$. 
These local patches are randomly chosen based on weighted sampling of foreground pixels using the ground truth image mask. 
Secondly, we downsample the original $256 \times 256$ images to smaller global images with a resolution of $128 \times 128$. 
Similar to the approach in LRM \cite{hong2023lrm}, we utilize the deferred back-propagation technique \cite{zhang2022arf} to conserve GPU memory. 
In essence, the adjusted loss function is as delineated below:
% In default, we set the resolution of rendered image as $256 \times 256$.
% However, direct computation on the whole $256 \times 256$ renderings with $L_{\text{overall}}$ is likely to encounter the overflow of GPU memory during training due to the huge memory consumption of NeRF.
% Consequently, we further adopt a simple yet effective solution to remedy this huge requirement on GPU memories by trading space with time.
% 1) We divide the original $256 \times 256$ images into smaller local patches with a resolution of $128 \times 128$.
% Given the mask of the ground truth image, the local patches are randomly selected with a weighted sampling on the foreground pixels.
% 2) We also downsample the original $256 \times 256$ images to smaller global images with a resolution of $128 \times 128$.
% Similar to LRM \cite{hong2023lrm}, we also use the deferred back-propagation technique \cite{zhang2022arf} to save GPU memory.
% In summary, the modified loss is as follows:
\begin{small}
    \begin{equation}
        L_{\text{overall}} = L_{\text{recon}}^{\text{local}} + L_{\text{recon}}^{\text{global}} + \lambda_{\text{adv}} L_{\text{adv}}^{\text{global}} + \lambda_{\text{clip}} L_{\text{clip}}^{\text{global}}
    \end{equation}
\end{small}
where $^{\text{global}}$ means the loss is computed on the global images. $^{\text{local}}$ means the loss is computed on the sampled local patches.

% \subsection{Training-free Multi-condition Generation}
% \label{method:controlrm:multi_condition}

\section{Experiment}
\label{sec-experiment}

% \subsection{Implementation Details}
\subsection{Experiment Details}

% \subsubsection{Training Data}
\subsubsection{Training Details}

Our training dataset comprises the training split of the G-Objaverse dataset \cite{qiu2023richdreamer}, which is a subset of Objaverse \cite{deitke2023objaverse}. 
We have randomly selected 260k samples from the original G-Objaverse for training, while the remaining samples are allocated for validation and evaluation purposes. 
The text prompts for each sample are sourced from Cap3D \cite{luo2024scalable}. 
Additionally, the visual condition maps are derived from the multi-view images in the dataset, encompassing edge, sketch, depth, and normal annotations. 
Edge annotations are generated using the Canny edge detector \cite{canny1986computational}, sketch annotations are produced with the sketch generation model from ControlNet \cite{zhang2023adding}.
Depth and normal annotations are provided by G-Objaverse, and further normalized to match the format of MVControl (Li et al., 2024).

% \subsubsection{Training Details}
We initialize our network using the weights from the pre-trained OpenLRM-base \cite{openlrm}. 
The image-conditioned transformer from OpenLRM is removed, and our proposed conditional backbone, incorporating text and visual conditions (such as sketch, edge, depth, and normal), is appended as input. 
During training, the cross-attention layers in the triplane transformer of OpenLRM are activated, while the remaining layers are kept frozen. 
We utilize the AdamW optimizer with a conservative learning rate of 4e-4 for training ControLRM on 16 Nvidia V100-32G GPUs. 
Each batch comprises 96 text-condition-image pairs. 
The training duration is estimated to be approximately 4-6 days for ControLRM-T and 5-6 days for ControLRM-D. 
The input resolution of the condition image is set to 336, while the rendered image resolution is set to 256

% We initialize our network by leveraging the weights of the pre-trained OpenLRM.
% We remove the image-conditioned transformer of OpenLRM and append our proposed conditional backbone with text and visual conditions (i.e. sketch/edge/depth/normal) as input.
% The cross-attention layers in the triplane transformer of OpenLRM are activated during training, while the remaining layers are frozen.
% We employ the AdamW optimizer with a conservative learning rate of 4e-4 to train ControLRM on 16 Nvidia V100-32G GPUs.
% Each batch contains 96 text-condition-image pairs in total.
% The training might take about 4-6 days for ControLRM-T and 5-6 days for ControLRM-D.
% The input resolution of the condition image is set to 336, and the rendered image resolution is set to 256.

\subsubsection{Evaluation Dataset}
\label{exp:details:evaluation_dataset}

For evaluation, we collect test samples from real world datasets rather than manually generated samples \cite{li2024controllable} to ensure unbiased generation.
Following the selection principle of MVControl \cite{li2024controllable} and TripoSR \cite{tochilkin2024triposr}, test data is gathered from three distinct datasets for comparative analysis in the subsequent experiments.

(1) \textbf{G-OBJ}: We collect 118 samples with highest clip scores between the text annotation and multi-view images from the test split of G-Objaverse dataset \cite{qiu2023richdreamer}, ensuring they are absent from the training data.
The text annotation is obtained from Cap3D \cite{luo2024scalable}.
We manually select one reference view from all provided 40 views in the dataset, and extract the edge/sketch/depth/normal condition maps on that reference view.
The remaining views are used as ground truth multi-view images for benchmark evaluation.

(2) \textbf{GSO}: We also collect 80 samples from the Google Scanned Objects dataset \cite{downs2022google} for zero-shot evaluation.
This dataset features more than one thousand 3D-scanned household items, serving as a valuable resource for assessing the zero-shot generalization capabilities of the proposed method.
In analogy with \textbf{G-OBJ}, we manually select a single reference view from the 32 available views in the dataset.
Subsequently, edge/sketch/depth/normal condition maps are generated for this chosen reference view.
Text annotations are obtained using BLIP2 \cite{li2023blip}.
The input data contains the prepared 2D condition map and the corresponding text prompt.
The remaining views are utilized as the ground truth for evaluation benchmark.

(3) \textbf{ABO}: We also select 80 samples from the Amazon Berkeley Objects dataset \cite{collins2022abo} for zero-shot evaluation. 
The Amazon Berkeley Objects dataset is a comprehensive 3D dataset comprising product catalog images, metadata, and artist-designed 3D models featuring intricate geometries and materials based on real household objects.
Text annotations are generated using BLIP2 caption model \cite{li2023blip}.
We manually select one reference view from the 72 available views provided in the dataset and extract the four condition maps (edge/sketch/depth/normal).
The remaining views are emplyed as ground truth for benchmark evaluation.

\subsubsection{Baselines}
\label{exp:details:baselines}

We compare our proposed ControLRM with other state-of-the-art baselines in the 3D generation task, including:
(1) \textbf{Score-Distillation-Sampling (SDS) methods}: GSGEN, GaussianDreamer, and DreamGaussians \cite{poole2022dreamfusion}; 
(2) \textbf{3D-based Diffusion models}: VolumeDiffusion and 3DTopia \cite{tang2023volumediffusion, hong20243dtopia}; 
(3) \textbf{Controllable 3D Diffusion models}: MVControl \cite{li2024controllable}. 
It is important to note that MVControl is the most relevant state-of-the-art controllable 3D generation method. 
For comparison purposes, we utilize the official implementations of the aforementioned methods in the subsequent experiments.

% We compare our proposed ControLRM with other state-of-the-art baselines in the 3D generation task, including:
% 1) Score-Distillation-Sampling (SDS) \cite{poole2022dreamfusion} methods: GSGEN \cite{chen2024text}, GaussianDreamer \cite{yi2024gaussiandreamer}, and DreamGaussians \cite{tang2023dreamgaussian}.
% 2) 3D-based Diffusion models: VolumeDiffusion \cite{tang2023volumediffusion}, and 3DTopia \cite{hong20243dtopia}.
% 3) Controllable 3D Diffusion models: MVControl \cite{li2024controllable}.
% Note that MVControl is the most related state-of-the-art controllable 3D generation method.
% For comparison, we run the official implementation of the aforementioned methods in the following experiments.

\begin{table}[t]
\centering
\caption{Quantitative results of controllability under \textbf{Edge (Canny)} condition in comparison with other SOTA 3D generation methods on \textbf{G-OBJ}. $\uparrow$ denotes higher result is better, while $\downarrow$ means lower is better. We report the metrics of \textbf{C-PSNR}, \textbf{C-SSIM}, and \textbf{C-MSE} in the table. The best results are highlighted with \textbf{\underline{underline}}, and the second best ones are highlighted with \textbf{\uwave{wavy-line}}.}
\vspace{-0.4cm}
\label{tab:controllability:canny}
\begin{tabular}{l|ccc}
\toprule
                                                  & \multicolumn{3}{c}{\textbf{Edge (Canny)}}                                                          \\
\multirow{-2}{*}{\textbf{Methods}}                 & \textbf{C-PSNR $\uparrow$}                & \textbf{C-SSIM $\uparrow$}                & \textbf{C-MSE $\downarrow$}                  \\ \midrule
{ \textbf{GSGEN} \cite{chen2024text}}             & { 11.54} & { 0.768} & { 0.0807} \\
{ \textbf{GaussianDreamer} \cite{yi2024gaussiandreamer}}   & { 11.08} & { 0.755} & { 0.0866} \\
{ \textbf{DreamGaussians} \cite{tang2023dreamgaussian}}    & { 8.98}  & { 0.667} & { 0.1341} \\
{ \textbf{VolumeDiffusion} \cite{tang2023volumediffusion}}   & { 11.75} & { 0.803} & { 0.0773} \\
{ \textbf{3DTopia} \cite{hong20243dtopia}}           & { 8.78}  & { 0.692} & { 0.1430} \\
{ \textbf{MVControl \cite{li2024controllable}}} & { 10.14} & { 0.738} & { 0.1052} \\
{ \textbf{ControLRM-T (Ours)}}       & { \textbf{\uwave{16.14}}} & { \textbf{\underline{0.891}}} & { \textbf{\uwave{0.0349}}} \\
{ \textbf{ControLRM-D (Ours)}}       & { \textbf{\underline{16.17}}} & { \textbf{\uwave{0.886}}} & { \textbf{\underline{0.0314}}} \\ \bottomrule
\end{tabular}
\vspace{-0.4cm}
\end{table}

\begin{table}[t]
\centering
\caption{Quantitative results of Controllability under \textbf{Sketch} condition in comparison with other SOTA 3D generation methods on \textbf{G-OBJ}. $\uparrow$ denotes higher result is better, while $\downarrow$ means lower is better. We report the metrics of \textbf{S-PSNR}, \textbf{S-SSIM}, and \textbf{S-MSE} in the table. The best results are highlighted with \textbf{\underline{underline}}, and the second best ones are highlighted with \textbf{\uwave{wavy-line}}.}
\vspace{-0.4cm}
\label{tab:controllability:sketch}
\begin{tabular}{l|ccc}
\toprule
                                                  & \multicolumn{3}{c}{\textbf{Sketch}}                                                                               \\
\multirow{-2}{*}{\textbf{Methods}}                 & \textbf{S-PSNR $\uparrow$}                & \textbf{S-SSIM $\uparrow$}                                     & \textbf{S-MSE $\downarrow$}                  \\ \midrule
{ \textbf{GSGEN} \cite{chen2024text}}             & { 13.19} & { 0.7629}                     & { 0.0499} \\
{ \textbf{GaussianDreamer} \cite{yi2024gaussiandreamer}}   & { 13.23} & { 0.7934}                     & { 0.0503} \\
{ \textbf{DreamGaussians} \cite{tang2023dreamgaussian}}    & { 13.14} & { 0.7740}                     & { 0.0516} \\
{ \textbf{VolumeDiffusion} \cite{tang2023volumediffusion}}   & { 15.16} & { 0.8247}                     & { 0.0326} \\
{ \textbf{3DTopia} \cite{hong20243dtopia}}           & { 13.73} & { 0.7902}                     & { 0.0443} \\
{ \textbf{MVControl \cite{li2024controllable}}} & { 12.56} & {{ 0.7406}} & { 0.0603} \\
{ \textbf{ControLRM-T (Ours)}}       & { \textbf{\underline{18.02}}} & { \textbf{\underline{0.9084}}}                     & { \textbf{\underline{0.0189}}} \\
{ \textbf{ControLRM-D (Ours)}}       & { \textbf{\uwave{17.56}}} & { \textbf{\uwave{0.9000}}}                     & { \textbf{\uwave{0.0208}}} \\ \bottomrule
\end{tabular}
\vspace{-0.4cm}
\end{table}

\begin{table}[t]
\centering
\caption{Quantitative results of Controllability under \textbf{Depth} condition in comparison with other SOTA 3D generation methods on \textbf{G-OBJ}. $\downarrow$ denotes lower result is better. We report the metrics of \textbf{M-MSE}, \textbf{Z-MSE}, and \textbf{R-MSE} in the table. The best results are highlighted with \textbf{\underline{underline}}, and the second best ones are highlighted with \textbf{\uwave{wavy-line}}.}
\vspace{-0.4cm}
\label{tab:controllability:depth}
\begin{tabular}{l|ccc}
\toprule
{ }                                   & \multicolumn{3}{c}{{ \textbf{Depth}}}                                                                        \\
\multirow{-2}{*}{{ \textbf{Methods}}} & { \textbf{M-MSE $\downarrow$}} & { \textbf{Z-MSE $\downarrow$}} & { \textbf{R-MSE $\downarrow$}} \\ \midrule
{ \textbf{GSGEN} \cite{chen2024text}}                     & { 0.1504}            & { 0.1381}          & { 0.0425}             \\
{ \textbf{GaussianDreamer} \cite{yi2024gaussiandreamer}}           & { 0.1019}            & { 0.1271}          & { 0.0558}             \\
{ \textbf{DreamGaussians} \cite{tang2023dreamgaussian}}            & { 0.1035}            & { 0.1284}          & { 0.0435}             \\
{ \textbf{VolumeDiffusion} \cite{tang2023volumediffusion}}           & { 0.1615}            & { 0.1156}          & { 0.0444}             \\
{ \textbf{3DTopia} \cite{hong20243dtopia}}                   & { 0.1364}            & { 0.1374}          & { 0.0412}             \\
{ \textbf{MVControl \cite{li2024controllable}}}         & { 0.0692}            & { 0.0695}          & { 0.0655}             \\
{ \textbf{ControLRM-T (Ours)}}               & { \textbf{\uwave{0.0287}}}            & { \textbf{\uwave{0.0198}}}          & { \textbf{\uwave{0.0355}}}             \\
{ \textbf{ControLRM-D (Ours)}}               & { \textbf{\underline{0.0285}}}            & { \textbf{\underline{0.0174}}}          & { \textbf{\underline{0.0331}}}             \\ \bottomrule
\end{tabular}
\vspace{-0.4cm}
\end{table}

\begin{table}[t]
\centering
\caption{Quantitative results of Controllability under \textbf{Normal} condition in comparison with other SOTA 3D generation methods on \textbf{G-OBJ}. $\downarrow$ denotes lower result is better. We report the metrics of \textbf{NB-MSE}, and \textbf{DN-Consistency} in the table. The best results are highlighted with \textbf{\underline{underline}}, and the second best ones are highlighted with \textbf{\uwave{wavy-line}}.}
\vspace{-0.4cm}
\label{tab:controllability:normal}
\begin{tabular}{l|cc}
\toprule
{ }                                   & \multicolumn{2}{c}{{ \textbf{Normal}}}                                                \\
\multirow{-2}{*}{{ \textbf{Methods}}} & { \textbf{NB-MSE $\downarrow$}} & { \textbf{DN-Consistency $\downarrow$}} \\ \midrule
{ \textbf{GSGEN} \cite{chen2024text}}                     & { 0.0140}                & { 0.0412}                           \\
{ \textbf{GaussianDreamer} \cite{yi2024gaussiandreamer}}           & { 0.0133}                & { 0.0404}                           \\
{ \textbf{DreamGaussians} \cite{tang2023dreamgaussian}}            & { 0.0141}                & { 0.0372}                           \\
{ \textbf{VolumeDiffusion} \cite{tang2023volumediffusion}}           & { 0.0129}                & { 0.0468}                           \\
{ \textbf{3DTopia} \cite{hong20243dtopia}}                   & { 0.0240}                & { 0.0431}                           \\
{ \textbf{MVControl \cite{li2024controllable}}}         & { 0.0103}                & { 0.0421}                           \\
{ \textbf{ControLRM-T (Ours)}}               & { \textbf{\uwave{0.0038}}}                & { \textbf{\uwave{0.0216}}}                           \\
{ \textbf{ControLRM-D (Ours)}}                        & { \textbf{\underline{0.0034}}}                & { \textbf{\underline{0.0205}}}                           \\ \bottomrule
\end{tabular}
\vspace{-0.4cm}
\end{table}

% \subsection{Quantitative Results of 3D Controllability}
\subsection{Experiment Results of 3D Controllability}

\begin{figure*}[t]
\centering
\includegraphics[width=0.9\linewidth]{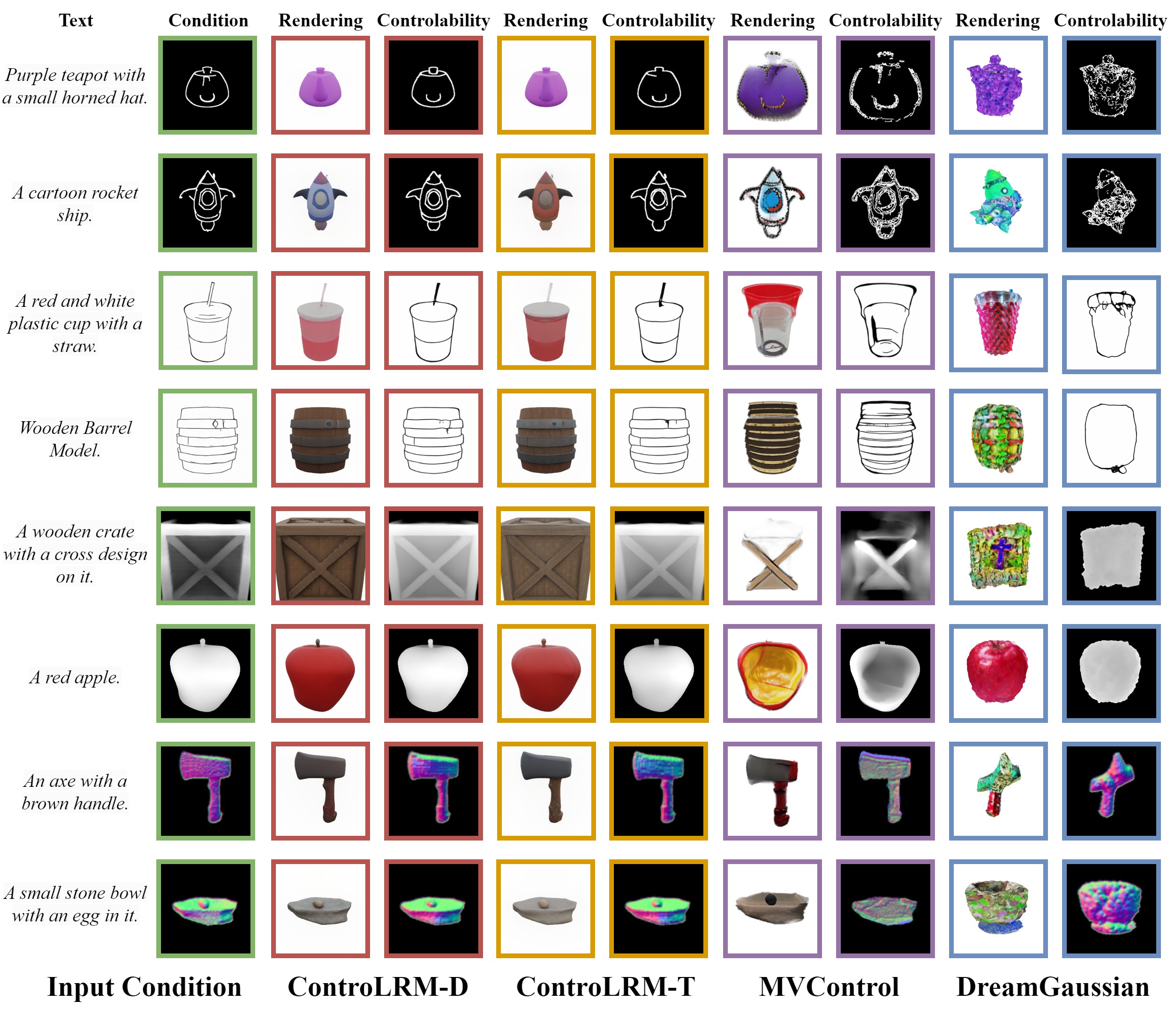}
\vspace{-0.4cm}
\caption{Visualization comparison of controllability under different conditional controls (Edge/Depth/Normal/Sketch).}
\vspace{-0.4cm}
\label{fig:comparison_controllability}
\end{figure*}

\subsubsection{Evluation Metrics}
\label{exp:controllability:metrics}

To assess the controllability of various 3D generation methods, we have developed metrics tailored to gauge the consistency of input 2D conditions following ControlNet++ \cite{li2024controlnet++}.
Four distinct conditions are taken into account: edge (canny), sketch, depth, and normal.
Specific metrics have been intricately designed for each condition to quantify the extent to which the condition is maintained throughout the generation process:

(1) \textbf{Edge Condition}:
Given the 2D edge map on the reference view, we use the generated 3D content to render a new image at the same view.
Subsequently, a Canny detector \cite{canny1986computational} is employed to extract the edge image from the rendered image, allowing for a comparison between the edge image and the original condition image.
The associated hyperparameters for Canny detector is the same as ControlNet \cite{zhang2023adding}.
To evaluate the resemblance of the edge maps, performance metrics such as Peak Signal-to-Noise Ratio (PSNR), Structural Similarity Index (SSIM), and Mean Squared Error (MSE) are computed following \cite{hong2023lrm}.
These metrics are further noted as \textbf{C-PSNR}, \textbf{C-SSIM}, and \textbf{C-MSE}.

(2) \textbf{Sketch Condition}:
Given the 2D sketch image on the reference view, we use the generated 3D content to render a new image at the same view.
Subsequently, the sketch extraction network provided by ControlNet \cite{zhang2023adding} is employed to derive the sketch map from the rendered image.
We emply PSNR, SSIM, MSE assess the similarity between the generated sketch map and the original sketch map.
These metrics are referred to as \textbf{S-PSNR}, \textbf{S-SSIM}, and \textbf{S-MSE} in this study.

(3) \textbf{Depth Condition}:
Given the 2D depth image on the reference view, we use the generated 3D content to render the image and the depth at the same viewpoint.

On the one hand, we can evaluate the depth consistency with foundation models in monocular depth estimation (i.e. Midas \cite{ranftl2020towards}, ZoeDepth \cite{bhat2023zoedepth}) following ControlNet++ \cite{li2024controlnet++}.
These foundation models are utilized to produce a depth map based on the input rendered image. Analogously, they are capable of estimating the depth map given a ground truth image as input on the reference view.
By leveraging the depth prior obtained from these foundation models, the Mean Squared Error (MSE) distance between the estimated depth maps of the ground truth image and the rendered image can indicate controllability under various depth conditions. 
When using Midas as the foundation model, the metric is denoted as \textbf{M-MSE}; whereas, if ZoeDepth is employed, the metric is referred to as \textbf{Z-MSE}.

On the other hand, an alternative method to assess depth consistency in 3D space involves comparing the disparity between the rendered depth map and the input conditional depth map.
The disparity, measured by MSE distance, between the rendered depth map and the input conditional depth map can also reflect the model's controllability performance.
However, discrepancies in scale between the estimated relative depth map and the input conditional depth map may adversely affect the accuracy of the MSE metric.
Thus, it becomes essential to address the scale discrepancy before evaluating the similarity between these depth maps.
Following the approach outlined in \cite{ranftl2020towards}, we compute an ordinary least squares solution to adjust for the scale and shift between these depth maps. 
Subsequently, the scale and shift transformation is applied to the relative depth map, and the MSE is then calculated between it and the input conditional depth map. 
This enables the calculation of a scale-agnostic MSE metric to evaluate the similarity between the depth maps, providing an effective way to evaluate the 3D consistency of the rendered depth map, denoted as \textbf{R-MSE}.

(4) \textbf{Normal Condition}:
Given the 2D normal map on the reference view, we use the generated 3D results to render the image and depth at the same viewpoint.

Firstly, we can assess the normal consistency with pre-trained models in surface normal estimation, such as Normal-BAE \cite{bae2021estimating} following ControlNet++ \cite{li2024controlnet++}.
The model for surface normal estimation facilitates the extraction of normal maps from rendered images. 
Similarly, the ground truth image can be input into the model to derive estimated normal maps. 
As the pre-trained model can grasp the surface normal priors from the input images, the Mean Squared Error (MSE) distance between these normal maps can indicate the controllability performance of the generation model. 
This evaluation metric, based on Normal-BAE, is referred to as \textbf{NB-MSE}.

Secondly, the evaluation of normal consistency in 3D space involves comparing the resemblance between the rendered depth maps and the input conditional normal maps.
The rendered depth map on the reference view is normalized to 0 to 1 first, and then used to calculate the normal map.
The MSE distance between this converted normal map and the input conditional normal map can demonstrate the normal consistency throughout the generation process. 
This metric, influenced by the depth-normal consistency in 3D space, is labeled as \textbf{DN-consistency}.

\begin{table*}[t]
\centering
\small
\caption{Quantitative comparison with SOTA 3d generation methods on G-Objaverse (\textbf{G-OBJ}) test set. We provide the zero-shot evaluation results of \textbf{FID $\downarrow$}, \textbf{CLIP-I $\uparrow$} and \textbf{CLIP-T $\uparrow$} on the test samples. The best results are highlighted with \textbf{\underline{underline}}, and the second best ones are highlighted with \textbf{\uwave{wavy-line}}. We also provide the time consumption of each method on a single V100-32G GPU to compare the efficiency.}
\vspace{-0.4cm}
\label{tab:comparison_generation:gobj}
\resizebox{\linewidth}{!}{
\begin{tabular}{l|c|ccc|ccc|ccc}
\toprule
\multirow{2}{*}{\diagbox{\textbf{Methods}}{\textbf{Metrics}}} & \multirow{2}{*}{\textbf{Time $\downarrow$}} & \multicolumn{3}{c|}{\textbf{Reference View}}                                                                          & \multicolumn{3}{c|}{\textbf{All Views}}                                                                               & \multicolumn{3}{c}{\textbf{Front-K Views}}                                                                     \\
         & & { \textbf{FID $\downarrow$}} & { \textbf{CLIP-I $\uparrow$}} & { \textbf{CLIP-T $\uparrow$}} & { \textbf{FID $\downarrow$}} & { \textbf{CLIP-I$\uparrow$}} & { \textbf{CLIP-T $\uparrow$}} & { \textbf{FID $\downarrow$}} & { \textbf{CLIP-I $\uparrow$}} & { \textbf{CLIP-T $\uparrow$}} \\ \midrule
{ \textbf{GSGEN} \cite{chen2024text}} & $\approx$ 40 min & { 235.09}       & { 0.756}           & { 0.308}           & { 340.39}       & { 0.762}           & { 0.298}           & { 357.13}       & { 0.781}           & { 0.310}           \\
{ \textbf{GaussianDreamer} \cite{yi2024gaussiandreamer}} & $\approx$ 2 min & { 182.70}       & { 0.802}           & { 0.295}           & { 268.95}       & { 0.803}           & { \textbf{\underline{0.309}}}           & { 282.59}       & { 0.823}           & { 0.321}           \\
{ \textbf{DreamGaussians} \cite{tang2023dreamgaussian}}  & $\approx$ 15 min & { 247.48}       & { 0.763}           & { 0.281}           & { 351.87}       & { 0.761}           & { 0.279}           & { 368.76}       & { 0.783}           & { 0.293}           \\
{ \textbf{VolumeDiffusion} \cite{tang2023volumediffusion}} & 142.55 sec & { 218.09}       & { 0.728}           & { 0.237}           & { 327.76}       & { 0.725}           & { 0.241}           & { 348.39}       & { 0.752}           & { 0.257}           \\
{ \textbf{3DTopia} \cite{hong20243dtopia}} & 177.89 sec
& { 228.79}       & { 0.719}           & { 0.267}           & { 289.02}       & { 0.749}           & { 0.280}           & { 329.29}       & { 0.808}           & { 0.312}           \\ 
{ \textbf{MVControl} \cite{li2024controllable}} & 8.92 sec & { 175.43}       & { 0.829}           & { 0.296}           & { 251.71}       & { 0.811}           & { 0.291}           & { 280.40}       & { 0.856}           & 0.318                         \\
{ \textbf{ControLRM-T (Ours)}} & 0.148 sec   & { \textbf{\underline{100.58}}}       & { \underline{\textbf{0.915}}}           & { \textbf{\uwave{0.309}}}           & { \textbf{\uwave{166.03}}}       & { \textbf{\uwave{0.879}}}           & { 0.292}           & { \underline{\textbf{144.02}}}       & { \textbf{\uwave{0.932}}}           & \textbf{\uwave{0.323}}                         \\
{ \textbf{ControLRM-D (Ours)}} & 0.503 sec   & { \textbf{\uwave{104.08}}}       & { \textbf{\uwave{0.911}}}           & { \underline{\textbf{0.315}}}           & { \textbf{\underline{163.25}}}       & { \textbf{\underline{0.887}}}           & { \textbf{\uwave{0.300}}}           & { \textbf{\uwave{148.76}}}       & { \textbf{\underline{0.935}}}           & { \textbf{\underline{0.330}}}                         \\ 
\bottomrule
\end{tabular}
}
\vspace{-0.4cm}
\end{table*}

\begin{table*}[t]
\centering
\small
\caption{Quantitative comparison with SOTA controllable 3D text-to-3d method (MVControl \cite{li2024controllable}) on G-Objaverse (\textbf{G-OBJ}) \cite{qiu2023richdreamer} test set. 4 kinds of different visual condition types are utilized for comparison here, including \textbf{Edge}, \textbf{Depth}, \textbf{Normal}, and \textbf{Sketch}. We provide the zero-shot evaluation results of \textbf{FID $\downarrow$}, \textbf{CLIP-I $\uparrow$} and \textbf{CLIP-T $\uparrow$} on the test samples. The best results are highlighted with \textbf{\underline{underline}}, and the second best ones are highlighted with \textbf{\uwave{wavy-line}}.}
\vspace{-0.4cm}
\label{tab:comparison_mvcontrol:gobj}
\resizebox{\linewidth}{!}{
\begin{tabular}{c|c|cccc|cccc|cccc}
\toprule
{ } & { } & \multicolumn{4}{c|}{{ \textbf{Reference View}}} & \multicolumn{4}{c|}{{ \textbf{All Views}}} & \multicolumn{4}{c}{{ \textbf{Front-K Views}}} \\
\multirow{-2}{*}{{ \textbf{Metrics}}} & \multirow{-2}{*}{{ \textbf{Methods}}} & \multicolumn{1}{c}{{ \textbf{Edge}}} & \multicolumn{1}{c}{{ \textbf{Depth}}} & \multicolumn{1}{c}{{ \textbf{Normal}}} & \multicolumn{1}{c|}{{ \textbf{Sketch}}} & \multicolumn{1}{c}{{ \textbf{Edge}}} & \multicolumn{1}{c}{{ \textbf{Depth}}} & \multicolumn{1}{c}{{ \textbf{Normal}}} & \multicolumn{1}{c|}{{ \textbf{Sketch}}} & \multicolumn{1}{c}{{ \textbf{Edge}}} & \multicolumn{1}{c}{{ \textbf{Depth}}} & \multicolumn{1}{c}{{ \textbf{Normal}}} & { \textbf{Sketch}} \\ \midrule
{ }                          & { \textbf{MVControl} \cite{li2024controllable}}                 & { 226.01}                    & { 158.38}                    & { 144.59}                     & { 172.73}                      & { 300.10}                    & { 229.45}                    & { 215.27}                     & { 262.03}                      & { 328.99}                    & { 257.62}                    & { 244.51}                     & { 290.49} \\
{ }                          & { \textbf{ControLRM-T}}               & { \textbf{\uwave{99.51}}}                    & { \textbf{\underline{102.88}}}                    & { \textbf{\underline{97.49}}}                      & { \textbf{\underline{102.43}}}                      & { \textbf{\uwave{165.21}}}                    & { \textbf{\underline{165.49}}}                    & { \textbf{\uwave{163.11}}}                     & { \textbf{\uwave{170.33}}}                      & { \textbf{\uwave{141.54}}}                    & { \textbf{\underline{147.18}}}                    & { \textbf{\underline{140.73}}}                     & { \textbf{\underline{146.63}}} \\
\multirow{-3}{*}{{ \textbf{FID $\downarrow$}}}     & { \textbf{ControLRM-D}}               & { \textbf{\underline{98.45}}}                     & { \textbf{\uwave{109.20}}}                    & { \textbf{\uwave{103.09}}}                     & { \textbf{\uwave{105.57}}}                      & { \textbf{\underline{158.73}}}                    & { \textbf{\uwave{166.36}}}                    & { \textbf{\underline{161.62}}}                     & { \textbf{\underline{166.28}}}                      & { \textbf{\underline{139.02}}}                    & { \textbf{\uwave{156.91}}}                    & { \textbf{\uwave{148.17}}}                     & { \textbf{\uwave{150.95}}} \\ \midrule
{ }                          & { \textbf{MVControl} \cite{li2024controllable}}                 & { 0.771}                     & { 0.854}                     & { 0.866}                      & { 0.825}                       & { 0.768}                     & { 0.831}                     & { 0.840}                      & { 0.806}                       & { 0.816}                     & { 0.875}                     & { 0.883}                      & { 0.851}  \\
{ }                          & { \textbf{ControLRM-T}}               & { \textbf{\uwave{0.915}}}                     & { \textbf{\underline{0.914}}}                     & { \textbf{\underline{0.919}}}                      & { \textbf{\underline{0.912}}}                       & { \textbf{\uwave{0.879}}}                     & { \textbf{\uwave{0.881}}}                     & { \textbf{\uwave{0.881}}}                      & { \textbf{\uwave{0.876}}}                       & { \textbf{\uwave{0.933}}}                     & { \textbf{\underline{0.932}}}                     & { \textbf{\uwave{0.932}}}                      & { \textbf{\uwave{0.930}}}  \\
\multirow{-3}{*}{{ \textbf{CLIP-I $\uparrow$}}}  & { \textbf{ControLRM-D}}               & { \textbf{\underline{0.920}}}                     & { \textbf{\uwave{0.902}}}                     & { \textbf{\uwave{0.912}}}                      & { \textbf{\uwave{0.911}}}                       & { \textbf{\underline{0.889}}}                     & { \textbf{\underline{0.885}}}                     & { \textbf{\underline{0.888}}}                      & { \textbf{\underline{0.886}}}                       & { \textbf{\underline{0.939}}}                     & { \textbf{\uwave{0.931}}}                     & { \textbf{\underline{0.935}}}                      & { \textbf{\underline{0.935}}}  \\ \midrule
{ }                          & { \textbf{MVControl} \cite{li2024controllable}}                 & { 0.262}                     & { \textbf{\underline{0.311}}}                     & { 0.312}                      & { 0.300}                       & { 0.264}                     & { \textbf{\underline{0.302}}}                     & { \textbf{\underline{0.304}}}                      & { 0.291}                       & { 0.295}                     & { \textbf{\uwave{0.326}}}                     & { \textbf{\underline{0.330}}}                      & { 0.318}  \\
{ }                          & { \textbf{ControLRM-T}}               & { \textbf{\uwave{0.309}}}                     & { \textbf{\uwave{0.308}}}                     & { \textbf{\uwave{0.310}}}                      & { \textbf{\uwave{0.309}}}                       & { \textbf{\uwave{0.291}}}                     & { 0.292}                     & { 0.293}                      & { 0.290}                       & { \textbf{\uwave{0.322}}}                     & { 0.323}                     & { \textbf{\uwave{0.324}}}                      & { \textbf{\uwave{0.322}}}  \\
\multirow{-3}{*}{{ \textbf{CLIP-T} $\uparrow$}}  & { \textbf{ControLRM-D}}               & { \textbf{\underline{0.318}}}                     & { \textbf{\underline{0.311}}}                     & { \textbf{\underline{0.316}}}                      & { \textbf{\underline{0.315}}}                       & { \textbf{\underline{0.301}}}                     & { \textbf{\uwave{0.299}}}                     & { \textbf{\uwave{0.300}}}                      & { \textbf{\underline{0.299}}}                       & { \textbf{\underline{0.332}}}                     & { \textbf{\underline{0.327}}}                     & { \textbf{\underline{0.330}}}                      & { \textbf{\underline{0.329}}}  \\ 
\bottomrule
\end{tabular}
}
\vspace{-0.4cm}
\end{table*}

\begin{table*}[t]
\centering
\small
\caption{Quantitative comparison with SOTA 3d generation methods on Google Scanned Objects (\textbf{GSO}) test set. We provide the zero-shot evaluation results of \textbf{FID $\downarrow$}, \textbf{CLIP-I $\uparrow$} and \textbf{CLIP-T $\uparrow$} on the test samples. The best results are highlighted with \textbf{\underline{underline}}, and the second best ones are highlighted with \textbf{\uwave{wavy-line}}. We also provide the time consumption of each method on a single V100-32G GPU to compare the efficiency.}
\vspace{-0.4cm}
\label{tab:comparison_generation:gso}
\resizebox{\linewidth}{!}{
\begin{tabular}{l|c|ccc|ccc|ccc}
\toprule
\multirow{2}{*}{\diagbox{\textbf{Methods}}{\textbf{Metrics}}}   &                                 & \multicolumn{3}{c|}{\textbf{Reference View}}                                                                          & \multicolumn{3}{c|}{\textbf{All Views}}                                                                               & \multicolumn{3}{c}{\textbf{Front-K Views}}                                                                              \\
     & \multirow{-2}{*}{\textbf{Time $\downarrow$}} & { \textbf{FID $\downarrow$}} & { \textbf{CLIP-I $\uparrow$}} & { \textbf{CLIP-T $\uparrow$}} & { \textbf{FID $\downarrow$}} & { \textbf{CLIP-I $\uparrow$}} & { \textbf{CLIP-T $\uparrow$}} & { \textbf{FID $\downarrow$}} & { \textbf{CLIP-I $\uparrow$}} & { \textbf{CLIP-T $\uparrow$}} \\ \midrule
{ \textbf{GSGEN} \cite{chen2024text}}           & $\approx$ 40 min                & { 273.54}       & { 0.734}           & { 0.286}           & { 344.61}       & { 0.740}           & { 0.289}           & { 360.57}       & { 0.759}           & { 0.300}           \\
{ \textbf{GaussianDreamer} \cite{yi2024gaussiandreamer}} & $\approx$ 2 min                 & { 189.41}       & { 0.815}           & { 0.305}           & { 278.70}       & { 0.810}           & { \textbf{\underline{0.300}}}           & { 287.20}       & { 0.829}           & { 0.311}           \\
{ \textbf{DreamGaussians} \cite{tang2023dreamgaussian}}  & $\approx$ 15 min                & { 271.80}       & { 0.761}           & { 0.281}           & { 359.65}       & { 0.760}           & { 0.279}           & { 373.53}       & { 0.784}           & { 0.290}           \\
{ \textbf{VolumeDiffusion} \cite{tang2023volumediffusion}} & 142.55 sec                      & { 236.01}       & { 0.719}           & { 0.261}           & { 299.61}       & { 0.715}           & { 0.259}           & { 316.35}       & { 0.742}           & { 0.273}           \\
{ \textbf{3DTopia} \cite{hong20243dtopia}}         & 177.89 sec                      & { 274.99}       & { 0.698}           & { 0.274}           & { 331.39}       & { 0.727}           & { 0.283}           & { 369.27}       & { 0.799}           & { 0.311}           \\
{ \textbf{MVControl} \cite{li2024controllable}}       & 8.92 sec                        & { 194.97}       & { 0.848}           & { 0.298}           & { 278.08}       & { 0.816}           & { \textbf{\uwave{0.288}}}           & { 301.31}       & { 0.870}           & { 0.312}           \\
{ \textbf{ControLRM-T (Ours)}}     & 0.148 sec                       & { \textbf{\underline{165.44}}}       & { \textbf{\underline{0.899}}}           & { \textbf{\uwave{0.309}}}           & { \textbf{\uwave{260.75}}}       & { \textbf{\underline{0.846}}}           & { 0.289}           & { \textbf{\uwave{251.57}}}       & { \textbf{\underline{0.912}}}           & { \textbf{\uwave{0.316}}}           \\
{ \textbf{ControLRM-D (Ours)}}     & 0.503 sec                       & { \textbf{\uwave{162.28}}}       & { \textbf{\uwave{0.896}}}           & { \textbf{\underline{0.313}}}           & { \textbf{\underline{171.13}}}       & { \textbf{\uwave{0.838}}}           & { \textbf{\underline{0.302}}}           & { \textbf{\underline{247.06}}}       & { \textbf{\uwave{0.908}}}           & { \textbf{\underline{0.322}}}           \\ \bottomrule
\end{tabular}
}
\vspace{-0.4cm}
\end{table*}

\begin{table*}[t]
\centering
\small
\caption{Quantitative comparison with SOTA controllable 3D text-to-3d method (MVControl \cite{li2024controllable}) on Google Scanned Objects (\textbf{GSO}) \cite{downs2022google} test set. 4 kinds of different visual condition types are utilized for comparison here, including \textbf{Edge}, \textbf{Depth}, \textbf{Normal}, and \textbf{Sketch}. We provide the zero-shot evaluation results of \textbf{FID $\downarrow$}, \textbf{CLIP-I $\uparrow$} and \textbf{CLIP-T $\uparrow$} on the test samples. The best results are highlighted with \textbf{\underline{underline}}, and the second best ones are highlighted with \textbf{\uwave{wavy-line}}.}
\vspace{-0.4cm}
\label{tab:comparison_mvcontrol:gso}
\resizebox{\linewidth}{!}{
\begin{tabular}{c|c|cccc|cccc|cccc}
\toprule
{ }                          & { }                          & \multicolumn{4}{c|}{{ \textbf{Reference View}}}                                                                    & \multicolumn{4}{c|}{{ \textbf{All Views}}}                                                                         & \multicolumn{4}{c}{{ \textbf{Front-K Views}}}                                                                        \\
\multirow{-2}{*}{{ \textbf{Metrics}}} & \multirow{-2}{*}{{ \textbf{Methods}}} & { \textbf{Edge}}  & { \textbf{Depth}}  & { \textbf{Normal}} & { \textbf{Sketch}} & { \textbf{Edge}}  & { \textbf{Depth}}  & { \textbf{Normal}} & { \textbf{Sketch}} & { \textbf{Edge}}  & { \textbf{Depth}}  & { \textbf{Normal}} & { \textbf{Sketch}} \\ \midrule
{ }                          & { \textbf{MVControl} \cite{li2024controllable}}                 & { 219.89} & { 187.77} & { 179.74} & { 192.48} & { 311.91} & { \textbf{\uwave{256.08}}} & { \textbf{\uwave{255.68}}} & { 288.64} & { 342.59} & { 272.83} & { 270.02} & { 319.81} \\
{ }                          & { \textbf{ControLRM-T}}               & { \textbf{\uwave{156.60}}} & { \textbf{\uwave{167.11}}} & { \textbf{\uwave{173.60}}} & { \textbf{\uwave{164.44}}} & { \textbf{\uwave{253.59}}} & { 262.95} & { 269.62} & { \textbf{\uwave{256.84}}} & { \textbf{\uwave{245.12}}} & { \textbf{\uwave{253.76}}} & { \textbf{\uwave{257.62}}} & { \textbf{\uwave{249.79}}} \\
\multirow{-3}{*}{{ \textbf{FID $\downarrow$}}}     & { \textbf{ControLRM-D}}               & { \textbf{\underline{151.48}}} & { \textbf{\underline{165.90}}} & { \textbf{\underline{171.33}}} & { \textbf{\underline{160.42}}} & { \textbf{\underline{165.38}}} & { \textbf{\underline{174.29}}} & { \textbf{\underline{174.98}}} & { \textbf{\underline{169.88}}} & { \textbf{\underline{234.42}}} & { \textbf{\underline{253.02}}} & { \textbf{\underline{256.24}}} & { \textbf{\underline{244.55}}} \\ \midrule
{ }                          & { \textbf{MVControl} \cite{li2024controllable}}                 & { 0.782}  & { 0.877}  & { \textbf{\underline{0.890}}}  & { 0.843}  & { 0.762}  & { \textbf{\uwave{0.841}}}  & { \textbf{\underline{0.851}}}  & { 0.811}  & { 0.815}  & { 0.895}  & { \textbf{\underline{0.907}}}  & { 0.864}  \\
{ }                          & { \textbf{ControLRM-T}}               & { \textbf{\uwave{0.915}}}  & { \textbf{\uwave{0.896}}}  & { \textbf{\uwave{0.879}}}  & { \textbf{\uwave{0.904}}}  & { \textbf{\underline{0.855}}}  & { \textbf{\underline{0.842}}}  & { \textbf{\uwave{0.835}}}  & { \textbf{\underline{0.852}}}  & { \textbf{\uwave{0.923}}}  & { \textbf{\underline{0.910}}}  & { \textbf{\uwave{0.894}}}  & { \textbf{\underline{0.919}}}  \\
\multirow{-3}{*}{{ \textbf{CLIP-I $\uparrow$}}}  & { \textbf{ControLRM-D}}               & { \textbf{\underline{0.916}}}  & { \textbf{\underline{0.892}}}  & { 0.870}  & { \textbf{\underline{0.906}}}  & { \textbf{\uwave{0.854}}}  & { 0.826}  & { 0.820}  & { \textbf{\uwave{0.850}}}  & { \textbf{\underline{0.928}}}  & { \textbf{\uwave{0.901}}}  & { 0.885}  & { \textbf{\underline{0.919}}}  \\ \midrule
{ }                          & { \textbf{MVControl} \cite{li2024controllable}}                 & { 0.265}  & { \textbf{\underline{0.312}}}  & { \textbf{\underline{0.312}}}  & { 0.301}  & { 0.263}  & { \textbf{\uwave{0.298}}}  & { \textbf{\uwave{0.299}}}  & { \textbf{\uwave{0.291}}}  & { 0.290}  & { \textbf{\underline{0.321}}}  & { \textbf{\underline{0.324}}}  & { 0.314}  \\
{ }                          & { \textbf{ControLRM-T}}               & { \textbf{\uwave{0.311}}}  & { 0.306}  & { 0.301}  & { \textbf{\underline{0.318}}}  & { \textbf{\uwave{0.293}}}  & { 0.288}  & { 0.284}  & { \textbf{\uwave{0.291}}}  & { \textbf{\uwave{0.320}}}  & { 0.317}  & { 0.311}  & { \textbf{\uwave{0.317}}}  \\
\multirow{-3}{*}{{ \textbf{CLIP-T $\uparrow$}}}  & { \textbf{ControLRM-D}}               & { \textbf{\underline{0.316}}}  & { \textbf{\underline{0.312}}}  & { \textbf{\uwave{0.308}}}  & { \textbf{\uwave{0.314}}}  & { \textbf{\underline{0.304}}}  & { \textbf{\underline{0.301}}}  & { \textbf{\underline{0.300}}}  & { \textbf{\underline{0.303}}}  & { \textbf{\underline{0.326}}}  & { \textbf{\uwave{0.317}}}  & { \textbf{\uwave{0.316}}}  & { \textbf{\underline{0.323}}}  \\ \bottomrule
\end{tabular}
}
\vspace{-0.4cm}
\end{table*}

\begin{table*}[t]
\centering
\small
\caption{Quantitative comparison with SOTA 3d generation methods on Amazon Berkeley Objects (\textbf{ABO}) test set. We provide the zero-shot evaluation results of \textbf{FID $\downarrow$}, \textbf{CLIP-I $\uparrow$} and \textbf{CLIP-T $\uparrow$} on the test samples. The best results are highlighted with \textbf{\underline{underline}}, and the second best ones are highlighted with \textbf{\uwave{wavy-line}}. We also provide the time consumption of each method on a single V100-32G GPU to compare the efficiency.}
\vspace{-0.4cm}
\label{tab:comparison_generation:abo}
\resizebox{\linewidth}{!}{
\begin{tabular}{l|c|ccc|ccc|ccc}
\toprule
\multirow{2}{*}{\diagbox{\textbf{Methods}}{\textbf{Metrics}}}    &                                 & \multicolumn{3}{c|}{\textbf{Reference View}}                                                                          & \multicolumn{3}{c|}{\textbf{All Views}}                                                                               & \multicolumn{3}{c}{\textbf{Front-K Views}}                                                                              \\
     & \multirow{-2}{*}{\textbf{Time $\downarrow$}} & { \textbf{FID $\downarrow$}} & { \textbf{CLIP-I $\uparrow$}} & { \textbf{CLIP-T $\uparrow$}} & { \textbf{FID $\downarrow$}} & { \textbf{CLIP-I $\uparrow$}} & { \textbf{CLIP-T $\uparrow$}} & { \textbf{FID $\downarrow$}} & { \textbf{CLIP-I $\uparrow$}} & { \textbf{CLIP-T $\uparrow$}} \\ \midrule
{ \textbf{GSGEN} \cite{chen2024text}}           & $\approx$ 40 min                & { 304.49}       & { 0.664}           & { 0.257}           & { 366.47}       & { 0.669}           & { 0.259}           & { 376.56}       & { 0.691}           & { 0.272}           \\
{ \textbf{GaussianDreamer} \cite{yi2024gaussiandreamer}} & $\approx$ 2 min                 & { 148.70}       & { 0.820}           & { 0.297}           & { 225.38}       & { 0.787}           & { 0.277}           & { 226.56}       & { 0.831}           & { 0.306}           \\
{ \textbf{DreamGaussians} \cite{tang2023dreamgaussian}}  & $\approx$ 15 min                & { 340.64}       & { 0.729}           & { 0.248}           & { 392.95}       & { 0.723}           & { 0.247}           & { 406.35}       & { 0.750}           & { 0.290}           \\
{ \textbf{VolumeDiffusion} \cite{tang2023volumediffusion}} & 142.55 sec                      & { 288.28}       & { 0.698}           & { 0.247}           & { 350.46}       & { 0.679}           & { 0.242}           & { 372.49}       & { 0.715}           & { 0.262}           \\
{ \textbf{3DTopia} \cite{hong20243dtopia}}         & 177.89 sec                      & { 247.89}       & { 0.692}           & { 0.273}           & { 231.55}       & { 0.751}           & { 0.272}           & { 259.88}       & { 0.844}           & { 0.312}           \\
{ \textbf{MVControl} \cite{li2024controllable}}       & 8.92 sec                        & { 149.61}       & { 0.857}           & { 0.305}           & { 217.97}       & { 0.802}           & { \textbf{\uwave{0.291}}}           & { 236.81}       & { 0.868}           & { 0.316}           \\
{ \textbf{ControLRM-T}}     & 0.148 sec                       & { \textbf{\uwave{85.08}}}        & { \textbf{\uwave{0.913}}}           & { \textbf{\uwave{0.311}}}           & { \textbf{\uwave{202.14}}}       & { \textbf{\underline{0.827}}}           & { 0.282}           & { \textbf{\uwave{160.12}}}       & { \textbf{\uwave{0.915}}}           & { \textbf{\uwave{0.320}}}           \\
{ \textbf{ControLRM-D}}     & 0.503 sec                       & { \textbf{\underline{80.12}}}        & { \textbf{\underline{0.914}}}           & { \textbf{\underline{0.320}}}           & { \textbf{\underline{181.84}}}       & { \textbf{\uwave{0.836}}}           & { \textbf{\underline{0.292}}}           & { \textbf{\underline{152.37}}}       & { \textbf{\underline{0.918}}}           & { \textbf{\underline{0.324}}}           \\ \bottomrule
\end{tabular}
}
\vspace{-0.4cm}
\end{table*}

\begin{table*}[t]
\centering
\small
\caption{Quantitative comparison with SOTA controllable 3D text-to-3d method (MVControl \cite{li2024controllable}) on Amazon Berekely Objects (\textbf{ABO}) \cite{collins2022abo} test set. 4 kinds of different visual condition types are utilized for comparison here, including \textbf{Edge}, \textbf{Depth}, \textbf{Normal}, and \textbf{Sketch}. We provide the zero-shot evaluation results of \textbf{FID $\downarrow$}, \textbf{CLIP-I $\uparrow$} and \textbf{CLIP-T $\uparrow$} on the test samples. The best results are highlighted with \textbf{\underline{underline}}, and the second best ones are highlighted with \textbf{\uwave{wavy-line}}.}
\vspace{-0.4cm}
\label{tab:comparison_mvcontrol:abo}
\resizebox{\linewidth}{!}{
\begin{tabular}{c|c|cccc|cccc|cccc}
\toprule
{ }                                   & { }                                   & \multicolumn{4}{c|}{{ \textbf{Reference View}}}                                                                                              & \multicolumn{4}{c|}{{ \textbf{All Views}}}                                                                                                   & \multicolumn{4}{c}{{ \textbf{Front-K Views}}}                                                                                               \\
\multirow{-2}{*}{{ \textbf{Metrics}}} & \multirow{-2}{*}{{ \textbf{Methods}}} & { \textbf{Edge}} & { \textbf{Depth}} & { \textbf{Normal}} & { \textbf{Sketch}} & { \textbf{Edge}} & { \textbf{Depth}} & { \textbf{Normal}} & { \textbf{Sketch}} & { \textbf{Edge}} & { \textbf{Depth}} & { \textbf{Normal}} & { \textbf{Sketch}} \\ \midrule
{ }                                   & { \textbf{MVControl}}                 & { 204.89}         & { 127.76}         & { 101.67}          & { 164.11}          & { 254.40}          & { \textbf{\uwave{197.09}}}         & { \textbf{\underline{183.39}}}    & { 236.99}          & { 268.67}         & { 221.58}         & { 201.11}          & { 255.85}          \\
{ }                                   & { \textbf{ControLRM-T}}               & { \textbf{\uwave{88.22}}}          & { \textbf{\uwave{93.60}}}          & { \textbf{\underline{84.00}}}     & { \textbf{\uwave{74.49}}}           & { \textbf{\uwave{207.31}}}         & { 220.16}         & { 196.77}          & { \textbf{\uwave{184.30}}}          & { \textbf{\uwave{162.46}}}         & { \textbf{\uwave{171.38}}}         & { \textbf{\underline{157.38}}}                & { \textbf{\uwave{149.24}}}                \\
\multirow{-3}{*}{{ \textbf{FID $\downarrow$}}}     & { \textbf{ControLRM-D}}               & { \textbf{\underline{74.89}}}    & { \textbf{\underline{86.32}}}    & { \textbf{\uwave{86.82}}}           & { \textbf{\underline{72.45}}}     & { \textbf{\underline{173.57}}}   & { \textbf{\underline{189.60}}}   & { \textbf{\uwave{192.93}}}          & { \textbf{\underline{171.27}}}    & { \textbf{\underline{149.25}}}   & { \textbf{\underline{156.44}}}   & { \textbf{\underline{159.65}}}          & { \textbf{\underline{144.15}}}                \\ \midrule
{ }                                   & { \textbf{MVControl}}                 & { 0.818}          & { 0.884}          & { 0.895}           & { 0.833}           & { 0.784}          & { 0.812}          & { 0.815}           & { 0.798}           & { 0.839}          & { 0.886}          & { 0.897}           & { 0.848}           \\
{ }                                   & { \textbf{ControLRM-T}}               & { \textbf{\uwave{0.909}}}          & { \textbf{\underline{0.909}}}    & { \textbf{\underline{0.907}}}     & { \textbf{\uwave{0.925}}}           & { \textbf{\uwave{0.828}}}          & { \textbf{\uwave{0.800}}}          & { \textbf{\underline{0.830}}}     & { \textbf{\uwave{0.848}}}           & { \textbf{\uwave{0.913}}}          & { \textbf{\uwave{0.911}}}          & { \textbf{\underline{0.908}}}     & { \textbf{\uwave{0.926}}}           \\
\multirow{-3}{*}{{ \textbf{CLIP-I $\uparrow$}}}  & { \textbf{ControLRM-D}}               & { \textbf{\underline{0.919}}}    & { \textbf{\underline{0.909}}}    & { \textbf{\uwave{0.898}}}           & { \textbf{\underline{0.931}}}     & { \textbf{\underline{0.850}}}    & { \textbf{\underline{0.814}}}    & { \textbf{\uwave{0.821}}}           & { \textbf{\underline{0.859}}}     & { \textbf{\underline{0.922}}}    & { \textbf{\underline{0.913}}}    & { \textbf{\uwave{0.903}}}           & { \textbf{\underline{0.934}}}     \\ \midrule
{ }                                   & { \textbf{MVControl}}                 & { 0.292}          & { 0.312}          & { \textbf{\underline{0.316}}}     & { 0.299}           & { \textbf{\uwave{0.282}}}          & { \textbf{\underline{0.295}}}    & { \textbf{\underline{0.300}}}       & { 0.287}           & { 0.310}           & { \textbf{\uwave{0.319}}}          & { \textbf{\underline{0.323}}}     & { 0.311}           \\
{ }                                   & { \textbf{ControLRM-T}}               & { \textbf{\uwave{0.302}}}          & { \textbf{\uwave{0.313}}}          & { 0.313}           & { \textbf{\uwave{0.317}}}           & { 0.276}          & { 0.279}          & { 0.284}           & { \textbf{\uwave{0.290}}}           & { \textbf{\underline{0.323}}}    & { 0.317}          & { 0.317}           & { \textbf{\uwave{0.321}}}           \\
\multirow{-3}{*}{{ \textbf{CLIP-T $\uparrow$}}}  & { \textbf{ControLRM-D}}               & { \textbf{\underline{0.319}}}    & { \textbf{\underline{0.320}}}    & { \textbf{\underline{0.316}}}     & { \textbf{\underline{0.326}}}     & { \textbf{\underline{0.294}}}    & { \textbf{\uwave{0.289}}}          & { \textbf{\uwave{0.285}}}           & { \textbf{\underline{0.300}}}     & { \textbf{\underline{0.323}}}    & { \textbf{\underline{0.324}}}    & { \textbf{\uwave{0.319}}}           & { \textbf{\underline{0.330}}}    \\ \bottomrule
\end{tabular}
}
\vspace{-0.4cm}
\end{table*}

\subsubsection{Quantitative Results}
% \subsubsection{Results with Canny Condition}
% \label{exp:controllability:canny}

\textbf{Results with Canny Condition}:
The comparison of the controllability of 3D generation methods under the Edge (Canny) condition is presented in Table \ref{tab:controllability:canny}. 
The aforementioned metrics of \textbf{C-PSNR}, \textbf{C-SSIM}, and \textbf{C-MSE} in Sec. \ref{exp:controllability:metrics} are utilized for evaluating controllability.
Our approach establishes a new state-of-the-art benchmark, surpassing other methods by a significant margin.
Specifically, our ControLRM-D and ControLRM-T achieve C-PSNR scores of 16.17 and 16.14 respectively, exhibiting an improvement of approximately 6 compared to the baseline performance.
Similar improvements can also be witnesses in C-SSIM and C-MSE.

% \subsubsection{Results with Sketch Condition}
% \label{exp:controllability:sketch}
\noindent\textbf{Results with Sketch Condition}
Tab. \ref{tab:controllability:sketch} presents the state-of-the-art comparison for the controllability of 3D generation results on Sketch condition.
The evaluation includes the results of three metrics introduced in Sec. \ref{exp:controllability:metrics}: \textbf{S-PSNR}, \textbf{S-SSIM}, and \textbf{S-MSE}.
These metrics can reflect how much sketch control information is preserved in the generated 3D results.
The results reveals that our models, ControLRM-D and ControLRM-T, outperform other methods significantly across all three metrics.
In comparison to the baseline method, MVControl, our approach showcases a significant enhancement, boasting around 6 points in S-PSNR, 0.25 in S-SSIM, and 0.04 in S-MSE.

% Tab. \ref{tab:controllability:sketch} presents the state-of-the-art comparison for the controllability of 3D generation results on Sketch condition.
% For evaluation, we report the results of \textbf{S-PSNR}, \textbf{S-SSIM}, and \textbf{S-MSE} discussed in Sec. \ref{exp:controllability:metrics}.
% These metrics can reflect how much control information of sketch is preserved in the generated 3D results.
% From the table, it can be found that the performance is much better than other methods on all 2 metrics of S-PSNR, S-SSIM and S-MSE.
% Our ControLRM-D and ControLRM-T respectively achieve S-PSNR of 17.56/18.02, S-SSIM of 0.9000/0.9084, S-MSE of 0.0208/0.0189.
% ControLRM-T with visual transformer backbone performs better than ControLRM-D on the controllability of Sketch condition.
% Compared with the baseline controllable 3D generation method (MVControl), our method can achieve a clear improvement of about 6 S-PSNR, 0.25 S-SSIM, and 0.04 S-MSE.

% \subsubsection{Results with Depth Condition}
% \label{exp:controllability:depth}
\noindent\textbf{Results with Depth Condition}
Tab. \ref{tab:controllability:depth} shows the state-of-the-art comparison for the controllability of 3D generation methods on Depth condition.
We report the scores of \textbf{M-MSE}, \textbf{Z-MSE}, and \textbf{R-MSE} introduced in Sec. \ref{exp:controllability:metrics}.
From the results in the table, our proposed methods, ControLRM-D and ControLRM-T, outperform other baselines across all three metrics of depth controllability.
Specifically, our proposed method demonstrates an improvement of approximately 0.04 in the M-MSE, 0.05 in Z-MSE, 0.03 in R-MSE, compared to MVControl.

% Tab. \ref{tab:controllability:depth} shows the state-of-the-art comparison for the controllability of 3D generation methods on Depth condition.
% We report the scores of \textbf{M-MSE}, \textbf{Z-MSE}, and \textbf{R-MSE} introduced in Sec. \ref{exp:controllability:metrics}.
% To measure the controllability of depth condition in 3D generation, we evaluate the results in 2 ways:
% 1) Model-based metrics of \textbf{M-MSE} and \textbf{Z-MSE}: The depth prior of large foundation models like Midas and ZoeDepth are utilized to calculate the depth consistency in these metrics.
% 2) Rendering-based metric of \textbf{R-MSE}: It reflects the difference between the rendered depth map and the input depth condition, which concentrates more on the exact 3D consistency.
% From the table, it can be found that our proposed ControLRM-D and ControLRM-T are performing better than other baselines in all 3 metrics of depth controllability.
% In the metric of M-MSE, our proposed method has an improvement of about 0.04 compared with MVControl.
% Similarly, in the metric of Z-MSE, our method improves the score by 0.05 compared with MVControl.
% For 3D consistency, our method can still achieve an improvement of 0.03 on R-MSE, achieving better performance than MVControl.

% \subsubsection{Results with Normal Condition}
% \label{exp:controllability:normal}
\noindent\textbf{Results with Normal Condition}
Tab. \ref{tab:controllability:normal} shows the state-of-the-art comparison for the controllability of 3D generation methods on Normal condition.
The evaluation metrics include \textbf{NB-MSE} and \textbf{DN-Consistency} introduced in Sec. \ref{exp:controllability:metrics}.
From the comparison results, our proposed ControLRM-D/ControLRM-T models outperforms other baselines in both NB-MSE and DN-Consistency metrics.
Specifically, ControLRM-D and ControLRM-T achieve NB-MSE scores of 0.0034 and 0.0038, respectively, representing a notable improvement compared to MVControl (0.0103 NB-MSE).
Significant improvment of our models in DN-Consistency score can also be found in the table.

% Tab. \ref{tab:controllability:normal} shows the state-of-the-art comparison for the controllability of 3D generation methods on Normal condition.
% The evaluation metric is \textbf{NB-MSE} and \textbf{DN-Consistency} introduced in Sec. \ref{exp:controllability:metrics}.
% In analogy with Sec. \ref{exp:controllability:depth}, we also evaluate the results in 2 ways:
% 1) Model-based metric of \textbf{NB-MSE}: The normal prior of pretrained mode (Normal-BAE \cite{bae2021estimating}) is used to measure the normal consistency.
% 2) Rendering-based metrid of \textbf{DN-Consistency}: The rendered depth map of generated 3D contents is supposed to satisfy the depth normal consistency given the input normal condition.
% From the table, it can be found that our proposed ControLRM-D/ControLRM-T performs better on both of these 2 metrics.
% ControLRM-D and ControLRM-T obtain the NB-MSE scores of 0.0034 and 0.0038, which is significantly better than MVControl (0.0103 NB-MSE).
% The DN-Consistency scores of ControLRM-D and ControLRM-T are 0.0205 and 0.0216, outperforming MVControl with an improvement of about 0.02 DN-Consistency score.

\subsubsection{Qualitative Results}

Controllable 3D generation requires the persistence of input conditions as a crucial ability. 
The generated 3D contents should retain the control information of the input conditions.
For qualitative comparison of 3D controllability, we visualize the generated results and the extracted condition maps in Fig. \ref{fig:comparison_controllability}.
The first two columns display the visualization of text and 2D visual conditions.
Subsequent columns exhibit the visualization results of the rendered images and the extracted visual condition map from them.
The comparison encompasses several methods: our ControLRM-D (columns 3-4), ControLRM-T (columns 5-6), MVControl (columns 7-8) \cite{li2024controllable}, and DreamGaussian (columns 9-10) \cite{tang2023dreamgaussian}.
Each row of the figure corresponds to a specific control condition: Rows 1-2 (Edge), Rows 3-4 (Sketch), Rows 5-6 (Depth), and Rows 7-8 (Normal).
As shown in the figure, ControLRM-D and ControLRM-T can effectively preserve the control information in the generated 3D content.
For instance, in the first and second rows, the controllability results of MVControl and DreamGaussian under the Canny condition appear noticeably fuzzier compared to those of ControLRM-D/T.
It demonstrates our proposed method can effectively maintain the controllability during 3D generation, providing better scalability compared with existing methods.

\begin{figure*}[t]
\centering
\includegraphics[width=0.9\linewidth]{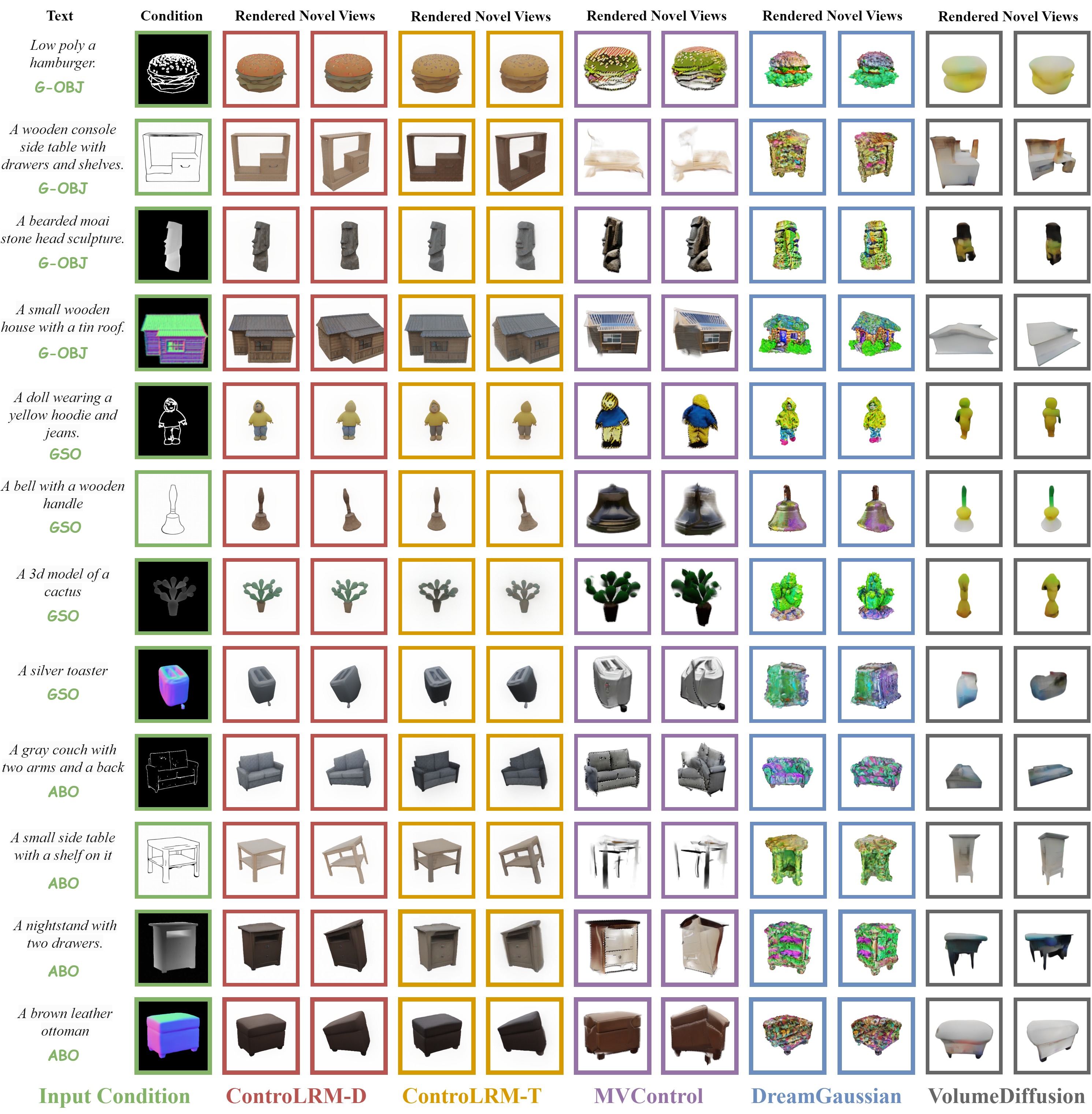}
\vspace{-0.4cm}
\caption{Qualitative comparison with SOTA 3D generation methods, including MVControl \cite{li2024controllable}, DreamGaussian \cite{tang2023dreamgaussian}, and VolumeDiffusion \cite{tang2023volumediffusion}. To avoid cherry-picking, the input conditions are extracted from \textbf{G-OBJ}, \textbf{GSO}, and \textbf{ABO} datasets. None of the images are observed by our model during training. Please zoom in for clearer visualization.}
\vspace{-0.4cm}
\label{fig:comparison_3dgen}
\end{figure*}

\begin{figure*}[t]
\centering
\includegraphics[width=0.9\linewidth]{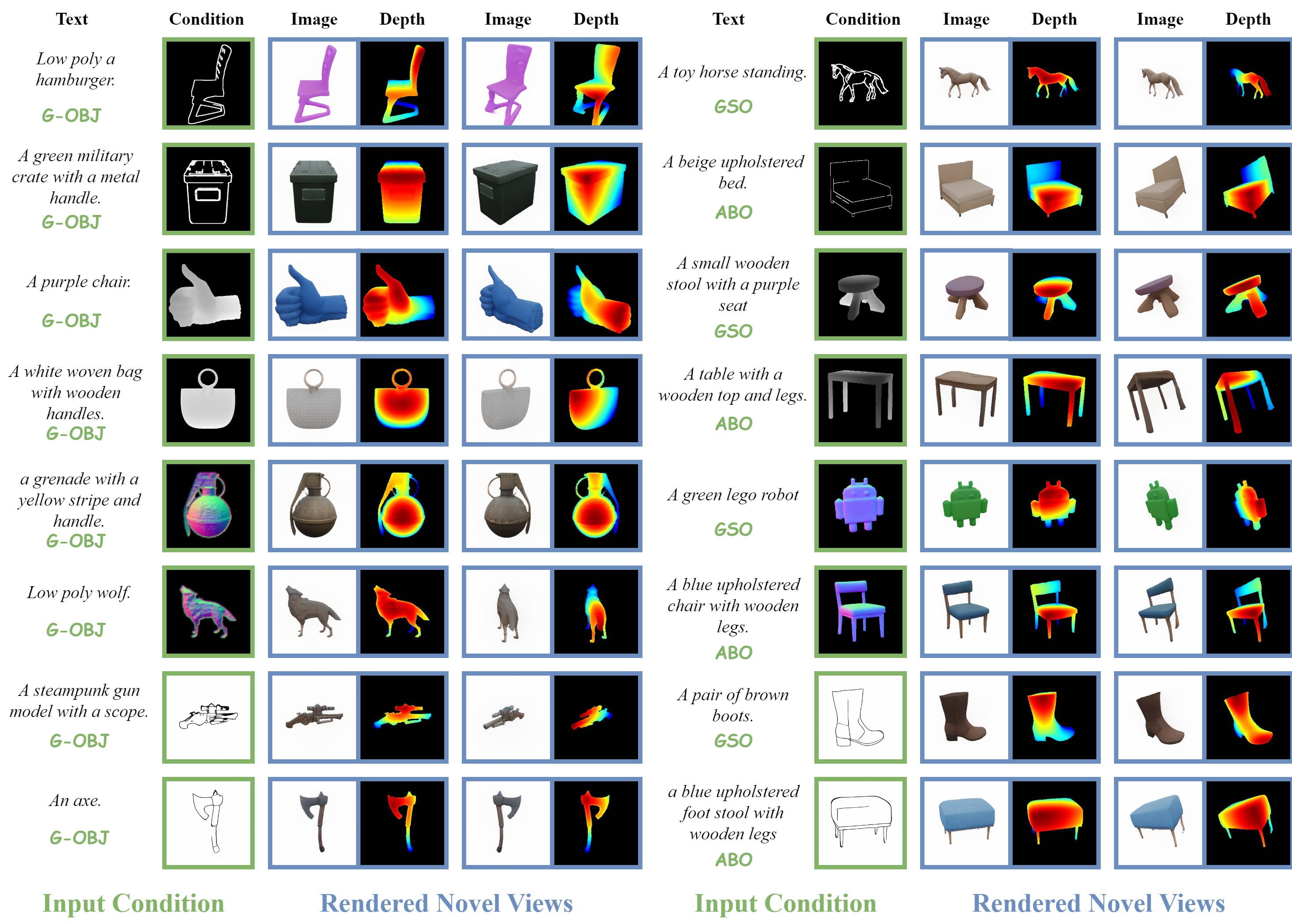}
\vspace{-0.4cm}
\caption{Visualization of rendered novel views (RGB and depth) generated by our ControLRM-D. The samples are extracted from \textbf{G-OBJ}, \textbf{GSO}, and \textbf{ABO} datasets. None of the images are observed by our model during training. Please zoom in for clearer visualization.}
\vspace{-0.4cm}
\label{fig:visualization_controlrmd}
\end{figure*}

\subsection{Experimental Results of Controllable 3D Generation}

\subsubsection{Evaluation Metrics}
\label{exp:generation:metrics}

For evaluation, we quantitatively compare our proposed method with baselines by measuring the quality of generated 3D contents with \textbf{FID}, the consistency to the reference ground truth image with \textbf{CLIP-I}, and the consistency to the reference text description with \textbf{CLIP-T}.

\textbf{Render FID}: Following LATTE3D \cite{xie2024latte3d}, we compute the Fréchet Inception Distance (FID) \cite{heusel2017gans} between the renderings of the generated 3D contents and the collected ground truth multi-view images.
This metric can measure how well the generated shapes align with those from the 2D prior in visual quality.

\textbf{CLIP-I}: Following MVControl \cite{li2024controllable}, we measure the CLIP scores of image features extracted from the renderings of the generated 3D contents and the collected ground truth images on different views.
This metric aims to reveal the similarity between the rendering results of generated 3D contents and the ground truth images.

\textbf{CLIP-T}: Following MVControl \cite{li2024controllable}, we also measure the CLIP scores of the image features extracted from the renderings and the given text prompt.
This metric can measure the similarity between the generated 3D contents and the given text descriptions.

\textbf{Multi-view Settings}: The evaluation protocol of MVControl \cite{li2024controllable} only calculate the CLIP score between the generated multi-view images and real ground truth images on the reference view.
However, merely evaluating the performance with ground truth on only one reference view is not comprehensive for comparing 3D generated contents.
Because a single view can only capture a portion of the 3D object, often omitting unseen parts.
Consequently, utilizing multi-view ground truth is essential to enhance the evaluation protocol.
As discussed in Sec. \ref{exp:details:evaluation_dataset}, we collect samples with multi-view ground truth from \textbf{G-OBJ}, \textbf{GSO}, and \textbf{ABO}.
By incorporating these multi-view samples, we enhance the original benchmark used in MVControl \cite{li2024controllable} to be more comprehensive in the following manner:
(1) \textbf{Reference View}: The rendered image and ground truth image on the reference view are utilized to compute metrics including FID, CLIP-I, and CLIP-T;
(2) \textbf{All Views}: All views are taken into account when calculating the three metrics between the rendered and ground truth images;
(3) \textbf{Front-K Views}: Given the provision of only one reference view, the views on the back side may lack crucial cues for precise prediction, potentially leading to unreliable results in multi-view scenarios. Therefore, incorporating an additional evaluation of the views in front of the reference view is necessary. Consequently, we select the K views closest to the given reference view for further metric computation, with the default value of K set to 4.

\subsubsection{Quantitative Comparison on G-OBJ}
\label{exp:generation:gobj}

To demonstrate the effectiveness of the proposed method in controllable 3D generation, we present the quantitative results on the \textbf{G-OBJ} benchmark in Tab. \ref{tab:comparison_generation:gobj} and \ref{tab:comparison_mvcontrol:gobj}.
Tab. \ref{tab:comparison_generation:gobj} shows the comparison of \textbf{FID}, \textbf{CLIP-I}, \textbf{CLIP-T} with other baselines.
We report the mean score of these metrics under four different conditions (edge/depth/normal/sketch).
The time efficiency of each method on a single V100-32G GPU is reported as well.
In the tables, we adopt three different multi-view settings during evaluation as discussed in Sec. \ref{exp:generation:metrics}.
As shown in Tab. \ref{tab:comparison_generation:gobj}, ControLRM-T achieves an inference speed of 0.148 seconds per sample, while ControLRM-D achieves 0.503 seconds per sample. 
Our ControLRM models significantly enhance the inference speed by an order of magnitude when compared to alternative methods.
In addition to the siginificant improvement in time efficiency, the benchmark results on nine metrics also show that our ControLRM can achieve significantly better performance than other baselines.
For example, ControLRM-D/ControLRM-T achieves 104.08/101.06 Reference FID score, 0.911/0916 Reference CLIP-I score, and 0.315/0.309 Reference CLIP-T score.
The baselines achieve over 175 FID score, which is significantly higher than ControLRM.
It demonstrates the superior ability and efficiency of the proposed method in controllable 3D generation.
Tab. \ref{tab:comparison_mvcontrol:gobj} shows the direct comparison with SOTA method (MVControl \cite{li2024controllable}) on four different visual conditions .
Similar to Tab. \ref{tab:comparison_generation:gobj}, the metrics of \textbf{FID}, \textbf{CLIP-I} and \textbf{CLIP-T} under three different multi-view settings are used to reveal the quality of the generated 3D contents.
On most of the evaluation metrics, our ControLRM can achieve competetive and even better performance than MVControl, and the inference speed is significantly faster
Specifically, the inference speeds of ControLRM-D (0.503 sec/sample) and ControLRM-T (0.148 sec/sample) are much faster than MVControl (8.92 sec/sample).
It demonstrates the superior ability and efficiency of the proposed method in controllable 3D generation.

\subsubsection{Quantitative Comparison on GSO}
\label{exp:generation:gso}

\begin{figure*}[t]
\centering
\includegraphics[width=0.9\linewidth]{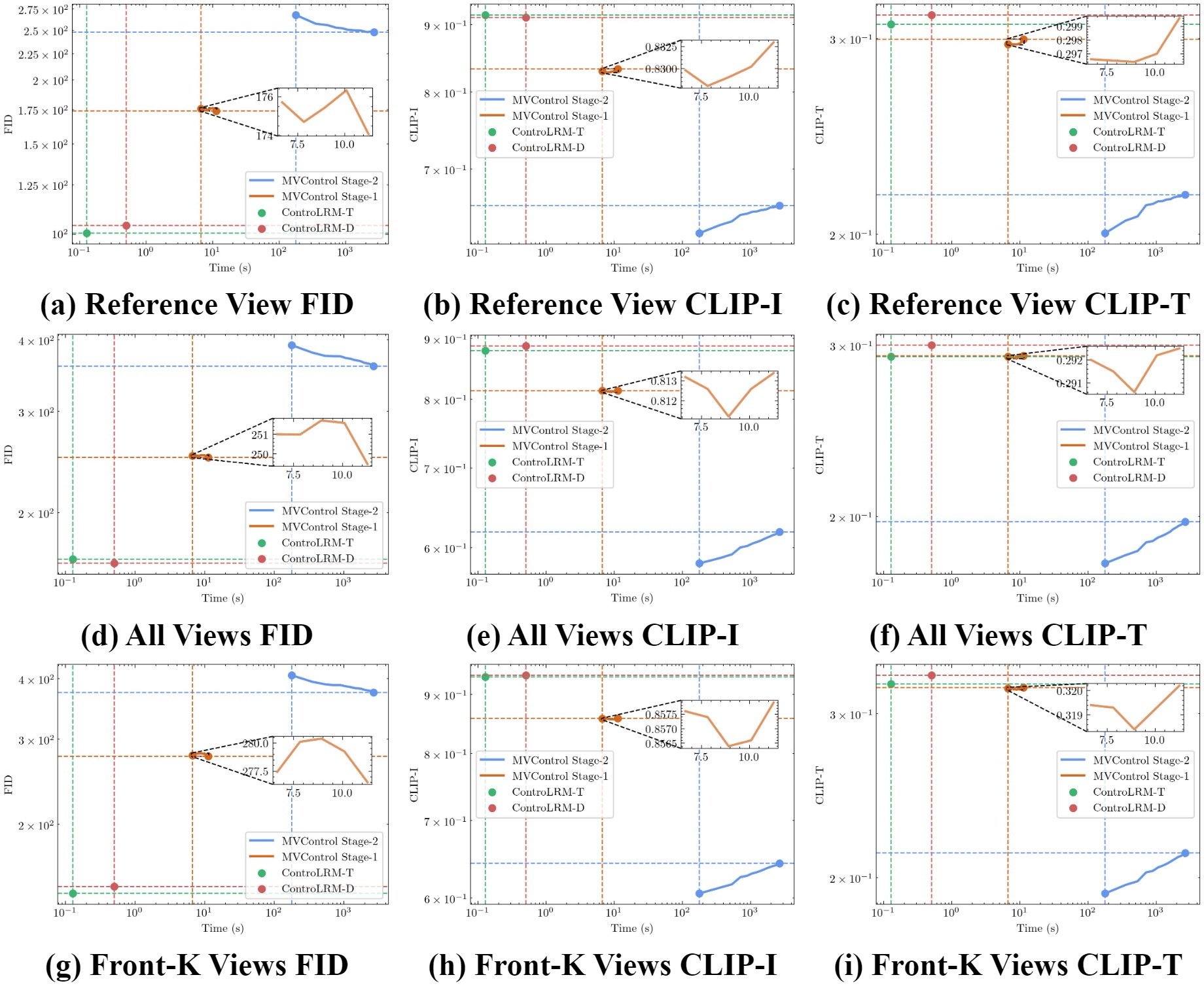}
\vspace{-0.4cm}
\caption{Visualization of the evaluation results (\textbf{FID}/\textbf{CLIP-I}/\textbf{CLIP-T}) at different amounts of optimization time on a single V100-32G GPU.
In comparison with the state-of-the-art controllable 3D generation method, MVControl \cite{li2024controllable}, our ControLRM can achieve over faster speed and better performance.}
\vspace{-0.4cm}
\label{fig:efficiency_comparison}
\end{figure*}

To demonstrate the generalization ability of the proposed method on the task of controllable 3D generation, we provide the experimental results on \textbf{GSO} benchmark and compare our model with other state-of-the-art methods introduced in Sec. \ref{exp:details:baselines}.
Similar to Sec. \ref{exp:generation:gobj}, we also use the evaluation metrics of \textbf{FID}, \textbf{CLIP-I}, and \textbf{CLIP-T} to measure the performance on controllable 3D generation.
These metrics are also calculated under 3 different multi-view settings as introduced in Sec. \ref{exp:generation:metrics}.
In Tab. \ref{tab:comparison_generation:gso}, we present the quantitative comparison among our proposed ControLRM and the baselines.
In most of the reported metrics, our ControLRM can achieve competetive and even better performance compared with the baselines.
As the table shows, our ControLRM-D and ControLRM-T outperform other baselines on the metrics of FID and CLIP-I in all view settings.
For example, under the reference view setting, our ControLRM-D/T can achieve 169.73/169.69 FID, significantly lower than the best one of the baselines, MVControl (194.97 FID score).
% For example, under the reference view setting, our ControLRM-D achieves 169.73 FID score and ControLRM-T achieves 169.69 FID score, significantly lower than the best one of the baselines, MVControl (194.97 FID score).
The zero-shot experiments on \textbf{GSO} can demonstrate the great generalization ability of the proposed method on unseen test cases.
We also provide the quantitative comparison between our ControLRM and MVControl \cite{li2024controllable} in Tab. \ref{tab:comparison_mvcontrol:gso} under 4 different input conditions.
The table shows that our proposed ControLRM is still competitive compared with MVControl.
For Edge and Sketch condition, both of ControLRM-D and ControLRM-T achieves better performance than MVControl in terms of FID, CLIP-I, and CLIP-T.
For the Depth and Normal conditions, ControLRM-D competes effectively with MVControl, although ControLRM-T shows slightly inferior performance.
An important reason is the preciseness of the given depth or normal map in controllable 3D generation.
Our ControLRM is trained using the ground truth depth or normal map of the dataset, which provides absolutely precise geometric prior as conditional input.
Whereas in the \textbf{GSO} benchmark, we extract the depth and normal maps using the annotator provided by MVControl.
The estimated depth and normal maps generated by the models provided by MVControl lack precision, leading to significant deviations in the predicted results. This inaccuracy can be misleading for ControLRM, which relies on precise geometric conditions.

\subsubsection{Quantitative Comparison on ABO}
\label{exp:generation:abo}

To evaluate the zero-shot generalization performance on controllable 3D generation, we further conduct experiments on \textbf{ABO} benchmark.
The quantitative comparison with other state-of-the-art methods in 3D generation introduced in Sec. \ref{exp:details:baselines} is presented in Tab. \ref{tab:comparison_generation:abo}.
The table employs the metrics of \textbf{FID}, \textbf{CLIP-I}, and \textbf{CLIP-T} to evaluate the performance of controllable 3D generation. These metrics are computed under three distinct multi-view settings discussed in Section \ref{exp:generation:metrics}.
From the table, we can find that ControLRM-D outperforms other baselines on all metrics.
ControLRM-T achieves the second best performance in most of these metrics.
In Tab. \ref{tab:comparison_mvcontrol:abo}, we compare our ControLRM with MVControl quantitatively under 4 different input conditions.
In the \textbf{ABO} benchmark, ControLRM-D exhibits competitive or superior performance in terms of \textbf{FID}, \textbf{CLIP-I}, and \textbf{CLIP-T} when compared to MVControl across all four conditions.
Conversely, our lightweight model, ControLRM-T, performs slightly less effectively than MVControl under depth and normal conditions but excels in canny and sketch conditions.
As outlined in Sec. \ref{exp:generation:gso}, the extraction of depth and normal maps relies on pre-trained models supplied by MVControl. 
Notably, ControLRM is trained using ground truth depth and normal maps, which differ from the estimated maps provided by the pre-trained models. 
This distribution discrepancy between the ground truth and estimated maps adversely impacts the performance of ControLRM.

\subsubsection{Qualitative Results}

In Fig. \ref{fig:comparison_3dgen}, we compare our ControLRM-D/T with state-of-the-art 3D generation methods: MVControl \cite{li2024controllable}, DreamGaussian \cite{tang2023dreamgaussian}, and VolumeDiffusion \cite{tang2023volumediffusion}.
The figure displays rendered novel views under four different condition controls (Edge/Depth/Normal/Sketch).
Our model demonstrates superior performance compared to other baselines, exhibiting higher quality and consistency in the generated 3D contents.
To ensure unbiased evaluation, we adopt input samples collected from \textbf{G-OBJ}, \textbf{GSO}, and \textbf{ABO} which are unseen in the training dataset following LRM \cite{hong2023lrm}.
The figure illustrates the capability of our ControLRM-D/T to infer semantically plausible 3D content from a single-view input visual condition.
% The figure illustrate the capability of our ControLRM-D/T to infer semantically plausible 3D content from a single-view input visual condition, while the baseline methods may produce inaccurate geometry and peculiar textures.
Additionally, we showcase more examples of generated 3D content from input conditions generated by \textbf{G-OBJ}, \textbf{GSO}, and \textbf{ABO} in Figure \ref{fig:visualization_controlrmd}, produced by our ControLRM-D.
The rendered images and depth maps in novel views are jointly visualized. Our model adeptly captures the intricate geometry of diverse input conditions (such as hands, guns, axes, etc.), and maintains consistent texture generation across the outputs.
The fidelity to the input visual conditions in the generated results underscores the exceptional performance and generalization capabilities of our model.

\begin{table}[t]
\centering
\small
\caption{Ablation analysis of each component in the training losses.}
\vspace{-0.4cm}
\label{tab:ablation}
\resizebox{\linewidth}{!}{
\begin{tabular}{c|cccccc}
\toprule
\multirow{2}{*}{Models} & \multicolumn{6}{c}{\textbf{Canny}}  \\
                        & \textbf{PSNR $\uparrow$}   & \textbf{SSIM $\uparrow$}   & \textbf{LPIPS $\downarrow$}   & \textbf{FID $\downarrow$}    & \textbf{CLIP-I $\uparrow$}    & \textbf{CLIP-T $\uparrow$}   \\ \midrule
Basic Training ($L_{\text{recon}}$) & 18.810	& 0.8198 & 0.1723 & 105.752	& 0.9061 & 0.3031        \\
+Adv Loss ($L_{\text{adv}}$)        & 19.445	& 0.8303 & 0.1581 & 100.163	& 0.9145 & 0.3085       \\
+CLIP Loss ($L_{\text{clip}}$)      & 19.452	& 0.8306 & 0.1579 & 99.867 & 0.9147 & 0.3087      \\
+2D Auxiliary ($x_{\text{aux}}$)    & 19.454	& 0.8306 & 0.1579 & 99.512 & 0.9150	& 0.3091        \\ \midrule
\multirow{2}{*}{Models} & \multicolumn{6}{c}{\textbf{Depth}}  \\
                        & \textbf{PSNR $\uparrow$}   & \textbf{SSIM $\uparrow$}   & \textbf{LPIPS $\downarrow$}   & \textbf{FID $\downarrow$}    & \textbf{CLIP-I $\uparrow$}    & \textbf{CLIP-T $\uparrow$}   \\ \midrule
Basic Training ($L_{\text{recon}}$) & 19.476 & 0.8314 & 0.1578 & 106.049 & 0.9079 & 0.3036       \\
+Adv Loss ($L_{\text{adv}}$)        & 20.051 & 0.8414 & 0.1469 & 103.625 & 0.9127 & 0.3068      \\
+CLIP Loss ($L_{\text{clip}}$)      & 20.066 & 0.8416 & 0.1465 & 103.220 & 0.9131 & 0.3075    \\
+2D Auxiliary ($x_{\text{aux}}$)    & 20.070 & 0.8417 & 0.1464 & 102.875 & 0.9135 & 0.3078       \\ \midrule
\multirow{2}{*}{Models} & \multicolumn{6}{c}{\textbf{Normal}}  \\
                        & \textbf{PSNR $\uparrow$}   & \textbf{SSIM $\uparrow$}   & \textbf{LPIPS $\downarrow$}   & \textbf{FID $\downarrow$}    & \textbf{CLIP-I $\uparrow$}    & \textbf{CLIP-T $\uparrow$}   \\ \midrule
Basic Training ($L_{\text{recon}}$) & 19.425 & 0.8312 & 0.1618 & 102.247 & 0.9133 & 0.3033      \\
+Adv Loss ($L_{\text{adv}}$)        & 19.903 & 0.8371 & 0.1518 & 98.694 & 0.9168 & 0.3063      \\
+CLIP Loss ($L_{\text{clip}}$)      & 19.905 & 0.8374 & 0.1517 & 97.724 & 0.9180 & 0.3103       \\
+2D Auxiliary  ($x_{\text{aux}}$)   & 19.909 & 0.8375 & 0.1516 & 97.489 & 0.9189 & 0.3103       \\ \midrule
\multirow{2}{*}{Models} & \multicolumn{6}{c}{\textbf{Sketch}}  \\
                        & \textbf{PSNR $\uparrow$}   & \textbf{SSIM $\uparrow$}   & \textbf{LPIPS $\downarrow$}   & \textbf{FID $\downarrow$}    & \textbf{CLIP-I $\uparrow$}    & \textbf{CLIP-T $\uparrow$}   \\ \midrule
Basic Training ($L_{\text{recon}}$) & 18.910 & 0.8205 & 0.1703 & 109.158 & 0.9023 & 0.3048        \\
+Adv Loss ($L_{\text{adv}}$)        & 19.546 & 0.8315 & 0.1588 & 103.164 & 0.9113 & 0.3085       \\
+CLIP Loss ($L_{\text{clip}}$)      & 19.552 & 0.8318 & 0.1585 & 102.710 & 0.9118 & 0.3087        \\
+2D Auxiliary ($x_{\text{aux}}$)    & 19.554 & 0.832 & 0.1583 & 102.426 & 0.9121 & 0.309      \\ \bottomrule
\end{tabular}}
\vspace{-0.6cm}
\end{table}

% \subsection{Efficiency Comparison}
\subsection{Extra Experiments}

\noindent\textbf{Efficiency Comparison:}
To provide a direct comparison of efficicency, we compare our ControLRM-D/T with the SOTA controllable 3D generation model MVControl \cite{li2024controllable} in Fig. \ref{fig:efficiency_comparison}.
MVControl consists of two stages: the first stage generates a coarse 3D content, and the second stage attempts to optimize the 3D content with test-time optimization using SDS loss \cite{poole2022dreamfusion}.
The quality of the generated 3D content improves over prolonged test-time optimization.
Both of these stages are compared in the figure.
We present visualizations of three evaluation metrics (FID, CLIP-I, CLIP-T) across three different multi-view settings (Reference View, All Views, Front-K Views) alongside the corresponding time consumption.
The average time consumed for generating a single 3D content per sample on a V100-32G GPU is reported.
We find that the refinement stage of MVControl tends to return worse performance than the coarse stage on the real-world data rather than the manually generated data used in their paper.

\noindent\textbf{Ablation Study:}
We conduct additional experiments to comprehensively analyize the contributions of the key components in our ControLRM framework. 
The ablation results under four different conditions are provided in Tab. \ref{tab:ablation}. 
By default, we utilized ControLRM-T in the ablation experiments. 
For evaluation, we reported the metrics of \textbf{PSNR}, \textbf{SSIM}, \textbf{LPIPS}, \textbf{FID}, \textbf{CLIP-I}, and \textbf{CLIP-T} in the table following MVControl \cite{li2024controllable}.
In the table, "Basic Training" indicates that the model was solely trained with the reconstruction loss $L_{\text{recon}}$. 
"+Adv Loss" signifies the addition of adversarial loss $L_{\text{adv}}$ to the reconstruction loss $L_{\text{recon}}$. 
Similarly, "+CLIP Loss" indicates the incorporation of clip loss $L_{\text{clip}}$. 
"+2D Auxiliary" refers to the adoption of auxiliary supervision on $x_{\text{aux}}$. 
The results demonstrate that the basic training scheme could achieve relatively good performance and meaningful generation with the support of large-scale pre-training weights from LRM \cite{hong2023lrm}, achieving a PSNR of approximately 18-19 in each of the four different conditions. 
The inclusion of adversarial loss $L_{\text{adv}}$ can led to an improvement of 3 to 5 in FID. 
Furthermore, the addition of clip loss $L_{\text{clip}}$ and 2D auxiliary supervision $x_{\text{aux}}$ can slightly enhance the FID by about 0.5. 
Overall, the results in the table highlight the effectiveness of each component in our ControLRM framework in enhancing the performance of controllable 3D generation.

\section{Limitation}
\label{sec-limitation}

In this study, the quantitative and qualitative analysis prove the superiority of our proposed method, but we also realize that this work is still insufficient and discuss the following limitations:
% In this study, we employ quantitative and qualitative analysis to assess the controllability and quality of generation achieved by the proposed method. 
% However, we also realize that this work is still insufficient and discuss the following issues:
\textbf{(1) Condition Expansion:} 
While significant advancements have been made under four control conditions, it is crucial to extend this framework to encompass additional control conditions such as segmentations, pose, and others.
\textbf{(2) Generalization Bottleneck:} 
The bottleneck of the proposed method is attributed to the utilization of the pre-trained Large Reconstruction Model (LRM). 
Although the proposed approach effectively aligns the controllable 2D generator with the pre-trained triplane decoder, failures in the pre-trained LRM could result in the failure of our ControLRM. 
Therefore, enhancing the performance by employing a more robust backbone can address this issue.

% In this paper, we use quantitative and qualitative analysis to evaluate the controllability and the generation quality of the proposed method.
% However, we also realize that this work is still insufficient and discuss the following issues:

% \noindent\textbf{Condition Expansion:} While we have achieved great improvements under 4 control conditions (Canny/Depth/Normal/Sketch), the extension to more control conditions (i.e. Segmentations, Pose, and etc) is still an important future direction. 

% \noindent\textbf{Generalization Bottleneck:} The bottleneck of the proposed method lies in the utitlized pre-trained Large Reconstruction Model (LRM).
% Though the proposed method can effectively align the controllable 2D generator with the pre-trained triplane decoder, the failure of pre-trained LRM might lead to the failure of our ControLRM.
% Consequently, improving the performance with a larger backbone (pre-trained LRM) can help promote this issue.

\section{Conclusion}
\label{sec-conclusion}

This paper introduces ControLRM, a novel controllable 3D generation framework characterized by high speed and superior generation quality. 
Our model offers support for four different types of controls: Edge (Canny), Depth, Normal, and Sketch. 
The architecture comprises an end-to-end feed-forward network that includes a 2D condition encoder based on transformer or diffusion models and a 3D triplane decoder leveraging a pre-trained LRM, where only the cross-attention layers are active during training. 
Additionally, we introduce an joint training pipeline encompassing adversarial loss, clip loss, and reconstruction loss. 
To ensure fair evaluation, we collect unseen evaluation samples from three different datasets: G-OBJ, GSO, and ABO. 
The comprehensive quantitative and qualitative evaluation findings demonstrate that our model surpasses existing state-of-the-art methods and achieves generation speeds significantly faster by an order of magnitude.

% In this paper, we propose ControLRM, a novel controllable 3D generation framework with fast speed and superior generation quality.
% Our model can support 4 different kinds of controls, including: Canny (edge), Depth, Normal, and Sketch.
% We design an end-to-end feed-forward network architecture that is comprised of: (1) 2D condition encoder based on transformer or diffusion models; (2) 3D triplane decoder of pre-trained LRM where only the cross-attention layers are activated during training.
% We also propose an adversarial training pipeline that is comprised of adversarial loss, clip loss, and reconstruction loss.
% To avoid cherry-picking, we extract unseen evaluation samples from existing datasets: G-OBJ, GSO, and ABO.
% The quantitative and qualitative evaluation results show that our model can outperform existing state-of-the-art methods and achieves an order of magnitude faster speed in generation.

\ifCLASSOPTIONcaptionsoff
  \newpage
\fi

\bibliographystyle{IEEEtran}
\bibliography{ref}

\end{document}